\pdfoutput=1
\documentclass[11pt]{article}

\usepackage[hidelinks]{hyperref}
\usepackage{amsthm}

\usepackage{amsmath}
\usepackage{multirow}
\usepackage{amssymb} % 或 \usepackage{amsfonts}
\usepackage{booktabs}
\usepackage{xcolor}
\usepackage{amssymb} % 提供 \checkmark 和 \times
\usepackage{tabularray}
\usepackage{longtable}
\usepackage{booktabs}
\usepackage{listings}
\usepackage{pifont}  % 提供 \ding{51}（✓）和 \ding{55}（✗）
\usepackage{xcolor}  % 提供颜色支持
\usepackage[dvipsnames]{xcolor}
\usepackage{supertabular}
\usepackage{algorithm}
\usepackage{algpseudocode}

\usepackage[final]{acl}

\usepackage{times}
\usepackage{latexsym}

\usepackage[T1]{fontenc}
\usepackage[utf8]{inputenc}

\usepackage{microtype}

\usepackage{inconsolata}

\usepackage[colorinlistoftodos]{todonotes}

\usepackage{graphicx}

\newcommand{\MarGrad}{%
  \textcolor{red}{M}%
  \textcolor{red!80!orange}{A}%
  \textcolor{red!60!orange}{R}%
}

\title{%
\MarGrad\textcolor{red!40!orange}{S}: Unleashing the Power of \textcolor{red!40!orange}{S}peculative Decoding\\
via \textcolor{red}{M}\textcolor{red!80!orange}{a}\textcolor{red!60!orange}{r}gin-Aware Verification}

\author{
  Jingwei Song$^{1*}$,
  Xinyu Wang$^{2,7*}$,
  Hanbin Wang$^{3}$,
  Xiaoxuan Lei$^{2,6}$,\\
  \textbf{Bill Shi}$^{4\dagger}$,
  \textbf{Shixin Han}$^{5}$,
  \textbf{Eric Yang}$^{4}$,
  \textbf{Xiao-Wen Chang}$^{2\dagger}$,
  \textbf{Lynn Ai}$^{4}$\\[1mm]
  $^{1}$The University of Hong Kong,
  $^{2}$McGill University,
  $^{3}$Peking University,\\
  $^{4}$gradient,
  $^{5}$Alibaba Cloud,
  $^{6}$Mila,
  $^{7}$SimpleWay\\[2mm]
\small{
  \parbox{\linewidth}{\centering \ttfamily
    \href{mailto:u3638265@connect.hku.hk}{songjingwei@connect.hku.hk},
    \href{mailto:xinyu.wang5@mail.mcgill.ca}{xinyu.wang5@mail.mcgill.ca},\\
    \href{mailto:tianyu@gradient.network}{tianyu@gradient.network},
    \href{mailto:chang@cs.mcgill.ca}{chang@cs.mcgill.ca}
  }
}
}

\begin{document}
\maketitle
\begingroup
\renewcommand\thefootnote{}
\footnotetext{* Equal contributions.}
\footnotetext{$^{\dagger}$Corresponding authors.}
\endgroup
\begin{abstract}
Speculative Decoding (SD) accelerates autoregressive large language model (LLM) inference by decoupling generation and verification.
While recent methods improve draft quality by tightly coupling the drafter with the target model, the verification mechanism itself remains largely unchanged, relying on strict token-level rejection sampling.
In practice, modern LLMs frequently operate in low-margin regimes where the target model exhibits weak preference among top candidates.
In such cases, rejecting plausible runner-up tokens yields negligible information gain while incurring substantial rollback cost, leading to a fundamental inefficiency in verification.

We propose \textbf{Margin-Aware Speculative Verification}, a training-free and domain-agnostic verification strategy that adapts to the target model's local decisiveness.
Our method conditions verification on decision stability measured directly from the target logits and relaxes rejection only when strict verification provides minimal benefit.
Importantly, the approach modifies only the verification rule and is fully compatible with existing target-coupled speculative decoding frameworks.
Extensive experiments across model scales ranging from \textbf{8B to 235B} demonstrate that our method delivers consistent and significant inference speedups over state-of-the-art baselines while preserving generation quality across diverse benchmarks. \noindent\textbf{The code is available at} \url{https://github.com/5SSjw/MARS}.
\end{abstract}

\section{Introduction}

Autoregressive Large Language Models (LLMs) suffer from high inference latency due to memory-bandwidth constraints \citep{shazeer2019fast}.
Speculative Decoding (SD) addresses this by decoupling generation and verification:
a lightweight \textit{draft model} proposes a sequence of tokens, which are then
verified in parallel by the target model \citep{leviathan2023fast}.
State-of-the-art methods like Medusa \citep{cai2024medusa} and EAGLE 2 \citep{li2024eagle2fasterinferencelanguage} have significantly advanced this paradigm by integrating draft heads directly with target features, enabling high-fidelity drafting with minimal overhead.
\begin{figure*}[t] \centering
    \includegraphics[width=1\textwidth]{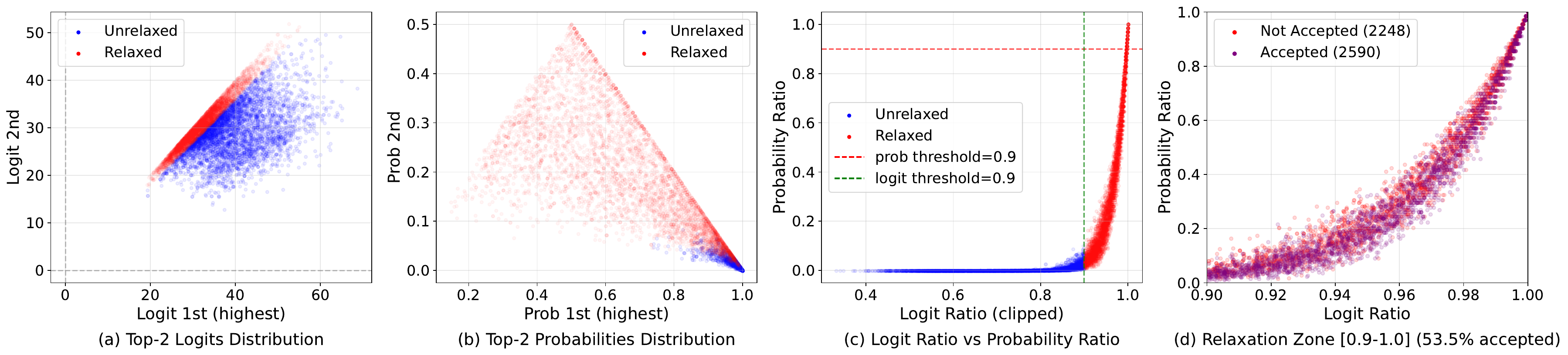}
    \caption{\textbf{Comparison of Logit Ratio vs. Probability Ratio for Adaptive Verification (Qwen3-8B).}
    Red points denote tokens accepted by our relaxation strategy (Logit Ratio $> 0.9$).
    \textbf{(a) Top-2 Logits:} Relaxed tokens cluster along the diagonal ($z_2 \approx z_1$), indicating our method captures candidates with similar raw scores regardless of scale.
    \textbf{(b) Top-2 Probabilities:} Unlike probability-based metrics, logit-based relaxation is not confined to high-entropy regions ($p_1 \approx p_2$); it also captures candidates with relatively lower probability ratios.
    \textbf{(c) Metric Decoupling:} High logit ratios (red) do not imply high probability ratios. The red points span the full range of probability ratios (y-axis), demonstrating that our method recovers valid candidates that are otherwise suppressed by the exponential sensitivity of the softmax function.
    \textbf{(d) Relaxation Zone Effectiveness:} A zoom-in on the low-margin regime (Logit Ratio $> 0.9$), where red points denote rejected tokens and purple points denote accepted ones.}
    \label{fig:scatter}
\end{figure*}
Despite these drafting improvements, the verification mechanism remains rigidly tied to exact-match rejection sampling.
This standard approach implicitly assumes the target model holds a decisive preference at every step.
In practice, however, LLMs frequently operate in \emph{low-margin regimes}, where the likelihood difference between top candidates is statistically negligible.
Strictly rejecting a plausible runner-up token in these cases yields negligible information gain while incurring substantial computational costs.
While some recent works explore relaxed verification via learned semantic judges \cite{bachmann2025judge,li2025loosely}, they often introduce complexity through supervised training and auxiliary models, limiting their plug-and-play capability.

In this work, we focus on the essence of the verification bottleneck: the mismatch between verification strictness and the target model's intrinsic uncertainty.
We propose a training-free paradigm that analyzes the target model's \emph{logit margin}—a direct proxy for decision stability.
We observe that wasteful rejections cluster precisely in low-margin regions. By conditioning verification on the stability of the target logits, we adaptively relax requirements when the model itself is indifferent.

We introduce \textbf{\underline{Mar}gin-Aware \underline{S}peculative Verification} (MARS), a simple strategy that performs stability-aware tie-breaking.
Our method modifies \emph{only} the verification rule, operating entirely at inference time without parameter updates.
Extensive experiments demonstrate that this lightweight principle unlocks the full potential of high-quality drafters.

\paragraph{Contributions.}
\begin{itemize}
    \item We identify \emph{low-margin rejections} as a fundamental inefficiency in strict verification, showing that adaptive tie-breaking can significantly reduce rollback costs.
    \item We propose a \textbf{training-free} verification algorithm MARS that leverages the logit margin to dynamically adapt verification rigor, avoiding the overhead of auxiliary judges.
    \item We demonstrate the scalability and robustness of MARS across a wide spectrum of model sizes (from \textbf{8B to 235B}) and diverse task settings. Extensive experiments confirm that our method consistently outperforms state-of-the-art baselines across chat, coding, and reasoning benchmarks, delivering significant acceleration with \textbf{near-lossless} accuracy preservation.
\end{itemize}
\section{Preliminaries}
\begin{figure*}[t]
  \centering
  \includegraphics[width=1\linewidth]{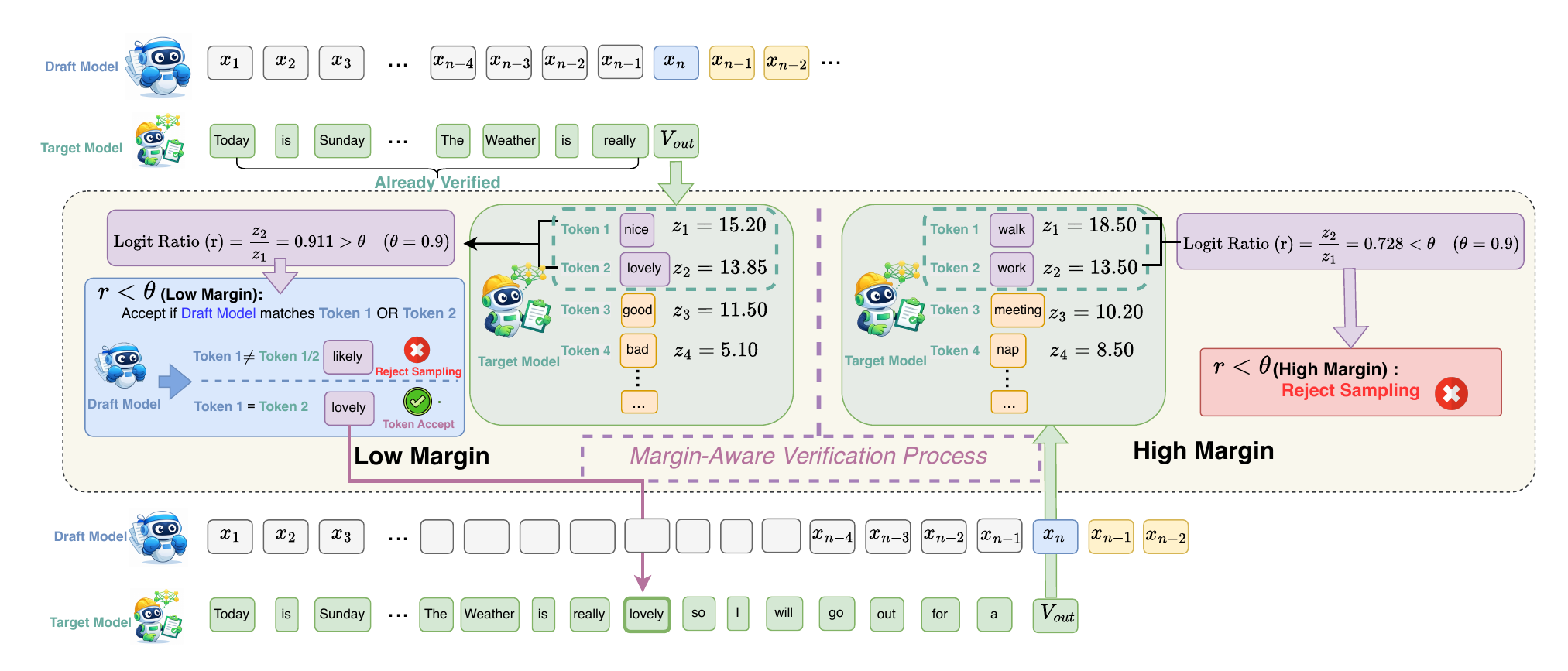} % Replace with your actual filename
 \caption{\textbf{Overview of the MARS Verification Workflow.} 
We illustrate the adaptive decision mechanism with a threshold $\theta=0.9$. 
\textbf{(Left) Low-Margin Regime:} The target model exhibits weak preference between the top candidates ``nice'' and ``lovely'' (Logit Ratio $r = 0.911 > \theta$). Identifying this local indifference, MARS accepts the draft token ``lovely'' (Top-2) as a valid tie-breaker, avoiding unnecessary rollback. 
\textbf{(Right) High-Margin Regime:} The target model decisively prefers ``walk'' over the runner-up ``work'' (Logit Ratio $r = 0.728 < \theta$). MARS detects this stability and reverts to strict verification, rejecting the draft to preserve generation fidelity.}
\label{fig:method_workflow}
\end{figure*}
\subsection{Autoregressive Generation}
Autoregressive Large Language Models (LLMs) generate text token-by-token.
Given a prefix sequence $x_{<t}$, the model computes the probability distribution
for the next token $x_t$ as $P(x_t \mid x_{<t})$.
This serial dependency prevents parallelization across time steps, making inference
memory-bandwidth bound on modern hardware (e.g., GPUs) and resulting in high latency.

\subsection{Speculative Decoding}
Speculative Decoding (SD) \citep{leviathan2023fast, chen2023speculative} alleviates
this bottleneck by employing a lightweight approximation model, referred to as the
\textit{draft model} $\mathcal{M}_s$.
At each step, $\mathcal{M}_s$ tentatively generates a short chain of $K$ candidate
tokens (the draft).
These tokens are then verified in parallel by the \textit{target model} $\mathcal{M}_t$ in a single forward pass, accepting those that match the target's distribution and discarding the rest.

Let $x_{<t}$ denote the current context.
The draft model $\mathcal{M}_s$ autoregressively generates a sequence of speculative tokens $\hat{V} = [\hat{v}_1, \dots, \hat{v}_K]$.
Subsequently, the target model $\mathcal{M}_t$ performs a parallel forward pass to
compute the conditional distributions $P(\cdot \mid x_{<t}, \hat{v}_{<i})$ for all
positions $i \in \{1, \dots, K\}$.
The core challenge lies in efficiently verifying these candidates while ensuring the final output distribution remains faithful to the target model $\mathcal{M}_t$.

\subsection{Speculative Verification}
\label{sec:spec_verify}

Speculative Decoding verifies draft tokens by comparing them against the target
model's output distribution.
In greedy decoding ($T{=}0$), a drafted token is accepted only if it matches the target's argmax prediction; the first mismatch truncates the draft chain and emits the target's top-1 token as correction.

In stochastic decoding ($T{>}0$), \citet{leviathan2023fast} propose a
rejection-sampling scheme that preserves the target distribution exactly.
Each drafted token $\hat{v}_i$ is accepted with probability
\begin{equation}
    \min\!\Bigl(1,\;\frac{P_{\mathcal{M}_t}(\hat{v}_i \mid x_{<t}, \hat{v}_{<i})}
    {P_{\mathcal{M}_s}(\hat{v}_i \mid x_{<t}, \hat{v}_{<i})}\Bigr),
    \label{eq:rejection_sampling}
\end{equation}
where $P_{\mathcal{M}_s}$ and $P_{\mathcal{M}_t}$ denote the draft and target
distributions. Upon rejection at position $i$, a corrected token is sampled
from a residual distribution that compensates for the draft's approximation
error. In both settings, \textbf{tree-based methods}
\citep{miao2024specinfer, li2024eagle2fasterinferencelanguage} extend this
logic by organizing candidates into a token tree, allowing the target model to
verify many competing prefixes in a single forward pass.

Our proposed method, MARS, modifies only the \emph{accept/reject decision}
during verification and is agnostic to the decoding strategy: it applies on
top of both greedy and sampling-based verification, as well as chain- and
tree-based draft structures.

\section{Methodology}
\label{sec:method}

\subsection{The Brittleness of Strict Verification}
\label{sec:motivation}
Speculative Decoding enforces strict token-level verification to guarantee
exact output matching with the target model.
This design implicitly treats the target model’s argmax prediction as a decisive
preference at every decoding step.
In practice, however, large language models frequently encounter
\emph{low-margin regimes}, where the likelihood difference between the top candidate tokens is small.
In such cases, the target model exhibits weak preference, and the choice of the  argmax token becomes arbitrary.
Rejecting a plausible draft token in these regimes provides negligible quality improvement, yet incurs a substantial computational cost due to rollback and resampling.
This mismatch between verification cost and benefit motivates a verification strategy that adapts to the target model’s local decisiveness, rather than enforcing rigid equality checks uniformly.

\subsection{Output distribution formulation}

Let $\mathcal{V}$ denote the vocabulary and $x_{<t}$ context at decoding step $t$.
An autoregressive language model computes a hidden representation
$\mathbf{h}_t \in \mathbb{R}^{d}$, which is projected to a logit vector
\begin{equation}
\mathbf{z}_t = \mathbf{W}_o \mathbf{h}_t + \mathbf{b}_o,
\end{equation}
where $\mathbf{W}_o \in \mathbb{R}^{|\mathcal{V}| \times d}$ and
$\mathbf{b}_o \in \mathbb{R}^{|\mathcal{V}|}$ are the output projection parameters.
We denote the component of $\mathbf{z}_t$ corresponding to a token $v$ as $z_{t,v}$. % Explicit definition
The conditional probability of $v \in \mathcal{V}$ is obtained via the softmax
transformation:
\begin{equation}
P(v \mid x_{<t}) = \frac{\exp(z_{t,v})}{\sum_{u \in \mathcal{V}} \exp(z_{t,u})}.
\end{equation}
Here, the logits $\mathbf{z}_t$ represent unnormalized scores that encode the
model's preference ranking over the vocabulary.

\subsection{Logit Ratio as a Confidence Margin Indicator}
\label{sec:ratio_indicator}
To quantify the model's confidence margin at token level, we consider the ratio between the raw logits $\mathbf{z}_t$ of the top 2 candidate tokens at step $t$. 
Let $v^{(1)}$ and $v^{(2)}$ denote the top-1 and top-2 token, and their corresponding logits are $z_{t, v^{(1)}}$ and $z_{t, v^{(2)}}$ such that $z_{t, v^{(1)}} \ge z_{t, v^{(2)}}$. For brevity, we denote these sorted logits as $z_{t, (1)}$ and $z_{t, (2)}$.

We propose the \textbf{Logit Ratio} ($r_t$) to characterize the confidence margin, instead of using standard probability-based metrics. Specifically, we compute:
\begin{equation}
\label{eq:ratio_def}
r_t = \frac{z_{t,(2)}}{z_{t,(1)}},
\end{equation}
A ratio approaching unity ($r_t \to 1$) implies that the gap between the best and second-best token is negligible, signaling \emph{low decisiveness}. We empirically examined the logit distributions of our target models (Figure \ref{fig:scatter}a) and observed that the top-ranked candidates consistently exhibit
positive logit values.
This \emph{positive domain dominance} ensures
that $r_t$ remains well-defined and strictly bounded in $(0, 1]$, where $r_t \to 1$
reliably signals low decisiveness.

We opt for logits because the softmax function introduces a non-linear distortion:
it exponentially magnifies the difference between candidates based on their absolute magnitude (Figure \ref{fig:scatter}b).

Specifically, the probability ratio $P(v^{(2)})/P(v^{(1)})$ is determined by the exponentiated difference of logits. 
Consequently, a fixed logit ratio ($z_{t,(2)}/z_{t,(1)}$) maps to vastly different probability ratios depending on the scale of $\mathbf{z}_t$ (Figure \ref{fig:scatter}c).
In regimes where logit values are large, the probability ratio becomes hypersensitive, often overestimating the model's actual confidence.
The logit ratio avoids this exponential scaling, providing a metric that is invariant to the global magnitude of the output vector.

% \[todo: add figures and captions and reference to figure 2\]

% Unlike probability-based metrics that require normalization or auxiliary models,
% $r_t$ is computed directly from raw logits and introduces no additional computational overhead.

To ground the ratio metric in standard decision theory, we relate it to the
logit margin, defined as $\Delta_t = z_{t,(1)} - z_{t,(2)}$.
Substituting $z_{t,(2)} = z_{t,(1)} - \Delta_t$ into Eq.~\ref{eq:ratio_def}, the
ratio can be rewritten as:
\begin{equation}
\label{eq:ratio_margin_relation}
r_t = 1 - \frac{\Delta_t}{z_{t,(1)}}.
\end{equation}
Consequently, enforcing the stability condition $r_t > \theta$ is equivalent to
imposing an \emph{adaptive margin constraint}:
\begin{equation}
\label{eq:adaptive_margin}
\Delta_t < (1 - \theta) \cdot z_{t,(1)}.
\end{equation}

Equation~\ref{eq:adaptive_margin} reveals a critical property: the required margin for stability is not fixed, but scales linearly with the magnitude of the top logit.
This counteracts the tendency of standard probability thresholds to overestimate confidence in high-logit regimes.
By effectively demanding a larger raw margin $\Delta_t$ when the model projects
larger values ($z_{t,(1)}$), the metric remains robust against calibration shifts that affect logit scale without altering the underlying preference ranking. Empirically, a substantial fraction of decoding steps fall into this low-margin region; Figure~\ref{fig:scatter}d visualizes the relaxation zone $r_t \in [0.9, 1.0]$, where the target model exhibits weak preference between the top candidates.

\subsection{Margin-Aware Speculative Verification Algorithm}

Building on the logit ratio analysis, we propose an \emph{Margin-Aware Speculative Verification Algorithm} that dynamically adjusts the acceptance rigor.
Unlike static verification, which enforces uniform exact matching, our method
modulates the strictness of the check based on the target model's local decisiveness.
This effectively balances the trade-off between \emph{generation quality} and
\emph{decoding latency}, relaxing constraints only when the target model exhibits negligible preference differences.

Given a draft token $\hat{v}_t$ proposed by the draft model $\mathcal{M}_s$, the verification workflow is illustrated in Figure~\ref{fig:method_workflow} and proceeds as follows:

\begin{itemize}
    \item \textbf{Exact Match.} If $\hat{v}_t = v^{(1)}$, the draft token matches the target model's primary prediction.
    It is accepted immediately, consistent with standard speculative decoding.

    \item \textbf{Adaptive Relaxation.} If $\hat{v}_t = v^{(2)}$ and $r_t > \theta$, the target model exhibits weak preference between the top two candidates.
    Rejecting the draft in this regime would incur a high rollback cost for
    negligible quality improvement.
    We therefore accept the draft token $\hat{v}_t$, effectively treating the
    prediction as a tie.

    \item \textbf{Rejection and Correction.} Otherwise, the draft token represents a significant deviation from the target distribution.
    The draft is rejected, and the target model's top choice $v^{(1)}$ is
    used as the correct token (standard rollback).
\end{itemize}

Throughout this work, we use a fixed threshold $\theta = 0.9$, a choice justified by extensive ablation studies in Section~\ref{ablation}. This policy effectively acts as a \textbf{cost-aware verification mechanism}:
strict verification is relaxed only in regimes where the target model is indecisive and where the expected quality improvement from rejection is negligible relative to the computational cost.
The complete procedure is summarized in Algorithm~\ref{alg:margin_verification}.

\begin{algorithm}[t]
\caption{Margin-Aware Speculative Verification Algorithm}
\label{alg:margin_verification}
\begin{algorithmic}[1]
\Require Draft $\hat{V} = [\hat{v}_1, \dots, \hat{v}_K]$, Target Model $\mathcal{M}_t$, Threshold $\theta$
\Ensure Verified sequence segment $V_{out}$

\State Initialize $V_{out} \leftarrow []$
\State Compute target logits $\mathbf{Z} = [\mathbf{z}_1, \dots, \mathbf{z}_K]$ in parallel

\For{$i = 1$ to $K$}
    \State Identify top-2 tokens $v^{(1)}, v^{(2)}$ from $\mathbf{z}_i$
    \State Compute ratio $r_i \leftarrow z_{i,(2)} / z_{i,(1)}$

    \If{$\hat{v}_i = v^{(1)}$}
        \State $V_{out}.\text{append}(\hat{v}_i)$ \Comment{Exact Match: Keep draft}
    \ElsIf{$\hat{v}_i = v^{(2)}$ \textbf{and} $r_i > \theta$}
        \State $V_{out}.\text{append}(\hat{v}_i)$ \Comment{Adaptive Relaxation: Keep draft}
    \Else
        \State $V_{out}.\text{append}(v^{(1)})$ \Comment{Correction: Use Target Top-1}
        \State \textbf{return} $V_{out}$ \Comment{Stop and discard remaining draft}
    \EndIf
\EndFor

\State \textbf{return} $V_{out}$ \Comment{All draft tokens accepted}
\end{algorithmic}
\end{algorithm}

% \paragraph{Choice of Verification Indicator.}
% Given the verification policy above, we briefly discuss the rationale for adopting
% the logit ratio over alternative indicators.
% A fixed margin threshold applies uniform tolerance across decoding steps, ignoring
% variations in logit scale across contexts and models.
% Probability-based ratios, while scale-invariant, require softmax normalization and do
% not induce the adaptive margin behavior described in Eq.~\ref{eq:adaptive_margin}.
% As shown in Section~\ref{sec:experiments}, the ratio-based indicator consistently
% achieves a superior speed--quality trade-off under the same verification policy.

\section{Experiments}
\label{sec:experiments}
% In this section, we describe the datasets, evaluation metrics, baselines, and implementation details.
% \usepackage{booktabs}
% \usepackage{multirow}

\begin{table*}[t]
\centering
\small
\setlength{\tabcolsep}{3pt}
\begin{tabular}{llcccccccccccc}
\toprule
& & \multicolumn{2}{c}{MT-bench} & \multicolumn{2}{c}{HumanEval} & \multicolumn{2}{c}{GSM8K} & \multicolumn{2}{c}{Alpaca} & \multicolumn{2}{c}{CNN/DM} & \multicolumn{2}{c}{Mean} \\
\cmidrule(lr){3-4}\cmidrule(lr){5-6}\cmidrule(lr){7-8}\cmidrule(lr){9-10}\cmidrule(lr){11-12}\cmidrule(lr){13-14}
Model & Method
& Speedup & $\tau$
& Speedup & $\tau$
& Speedup & $\tau$
& Speedup & $\tau$
& Speedup & $\tau$
& Speedup & $\tau$ \\
\midrule
% \multicolumn{14}{c}{Temperature=1} \\
% \midrule

\multirow{6}{*}{V 13B}
& SpS     & 1.62$\times$ & 1.84 & 1.72$\times$ & 1.97 & 1.46$\times$ & 1.73 & 1.52$\times$ & 1.78 & 1.66$\times$ & 1.89 & 1.60$\times$ & 1.84 \\
% & PLD       & 1.58$\times$ & 1.63 & 1.85$\times$ & 1.93 & 1.68$\times$ & 1.73 & 1.16$\times$ & 1.19 & 2.42$\times$ & 2.50 & 1.74$\times$ & 1.80 \\
% & Medusa    & 2.07$\times$ & 2.59 & 2.50$\times$ & 2.78 & 2.23$\times$ & 2.64 & 2.08$\times$ & 2.45 & 1.71$\times$ & 2.09 & 2.12$\times$ & 2.51 \\
% & Lookahead & 1.65$\times$ & 1.69 & 1.71$\times$ & 1.75 & 1.81$\times$ & 1.90 & 1.46$\times$ & 1.51 & 1.46$\times$ & 1.50 & 1.62$\times$ & 1.67 \\
& Lookahead & 1.45$\times$ & 1.39 & 1.41$\times$ & 1.25 & 1.61$\times$ & 1.60 & 1.36$\times$ & 1.36 & 1.26$\times$ & 1.20 & 1.42$\times$ & 1.36 \\
& PLD       & 1.38$\times$ & 1.33 & 1.55$\times$ & 1.43 & 1.48$\times$ & 1.43 & 1.06$\times$ & 1.04 & 2.22$\times$ & 2.20 & 1.54$\times$ & 1.49 \\
& Medusa    & 1.87$\times$ & 2.29 & 2.20$\times$ & 2.28 & 2.03$\times$ & 2.34 & 1.98$\times$ & 2.30 & 1.51$\times$ & 1.79 & 1.92$\times$ & 2.20 \\

& EAGLE   & 2.32$\times$ & 3.20 & 2.65$\times$ & 3.63 & 2.57$\times$ & 3.60 & 2.45$\times$ & 3.57 & 2.23$\times$ & 3.26 & 2.44$\times$ & 3.45 \\
& EAGLE-2 & 2.90$\times$ & 4.38 & 3.01$\times$ & 4.90 & 2.97$\times$ & 4.30 & 2.89$\times$ & 4.35 & 2.46$\times$ & 3.94 & 2.85$\times$ & 4.37 \\
& EAGLE-3    & 3.24$\times$ & 5.46 & 3.60$\times$ & 6.10 & 3.26$\times$ & 5.65 & 3.20$\times$ & 5.41 & 2.69$\times$ & 5.57 & 3.12$\times$ & 5.64 \\
& MARS      & \textbf{3.75$\times$} & \textbf{7.11} & \textbf{4.20$\times$} & \textbf{7.57} & \textbf{3.74$\times$} & \textbf{7.19} & \textbf{3.74$\times$} & \textbf{7.20} & \textbf{3.29$\times$} & \textbf{6.94} & \textbf{3.74$\times$} & \textbf{7.20} \\
\midrule

\multirow{2}{*}{L31 8B}
% & EAGLE-2   & 3.16$\times$ & 4.05 & 3.66$\times$ & 4.71 & 3.39$\times$ & 4.24 & 3.28$\times$ & 4.12 & 2.65$\times$ & 3.45 & 3.23$\times$ & 4.11 \\
& EAGLE-3    & 3.63$\times$ & 5.67 & 2.90$\times$ & 4.10 & 3.41$\times$ & 4.87 & 3.51$\times$ & 5.09 & 2.77$\times$ & 4.36 & 3.24$\times$ & 4.82 \\
& MARS      & \textbf{4.01$\times$} & \textbf{6.78} & \textbf{4.09$\times$} & \textbf{6.59} & \textbf{4.04$\times$} & \textbf{6.68} & \textbf{4.26$\times$} & \textbf{7.09} & \textbf{3.60$\times$} & \textbf{5.91} & \textbf{4.00$\times$} & \textbf{6.61} \\
\midrule

\multirow{2}{*}{L31 70B}
% & EAGLE-2   & 2.83$\times$ & 3.67 & 3.12$\times$ & 4.09 & 2.83$\times$ & 3.69 & 3.03$\times$ & 3.92 & 2.44$\times$ & 3.55 & 2.85$\times$ & 3.78 \\
& EAGLE-3    & 4.20$\times$ & 5.41 & 4.72$\times$ & 5.95 & 4.62$\times$ & 5.94 & 4.81$\times$ & 6.04 & 3.63$\times$ & 4.96 & 4.40$\times$ & 5.66 \\
& MARS       & \textbf{4.62$\times$} & \textbf{6.24} & \textbf{5.12$\times$} & \textbf{6.64} & \textbf{4.98$\times$} & \textbf{6.66} & \textbf{5.09$\times$} & \textbf{6.81} & \textbf{3.99$\times$} & \textbf{5.61} & \textbf{4.76$\times$} & \textbf{6.39} \\
\midrule

\multirow{2}{*}{Q3 8B}
& EAGLE-3    & 2.91$\times$ & 4.08 & 3.17$\times$ & 4.73 & 3.26$\times$ & 5.09 & 2.88$\times$ & 4.42 & 2.56$\times$ & 3.75 & 2.96$\times$ & 4.41 \\
& MARS      & \textbf{3.14$\times$} & \textbf{5.06} & \textbf{3.44$\times$} & \textbf{5.32} & \textbf{3.69$\times$} & \textbf{5.84} & \textbf{3.17$\times$} & \textbf{5.47} & \textbf{2.70$\times$} & \textbf{4.39} & \textbf{3.23$\times$} & \textbf{5.22} \\
\midrule

\multirow{2}{*}{Q3 32B}
& EAGLE-3    & 2.62$\times$ & 3.63 & 3.30$\times$ & 4.78 & 3.53$\times$ & 5.12 & 2.79$\times$ & 3.56 & 2.82$\times$ & 3.97 & 3.01$\times$ & 4.21 \\
& MARS      & \textbf{3.18$\times$} & \textbf{4.79} & \textbf{3.73$\times$} & \textbf{5.63} & \textbf{3.94$\times$} & \textbf{6.04} & \textbf{3.34$\times$} & \textbf{4.76} & \textbf{3.28$\times$} & \textbf{4.93} & \textbf{3.49$\times$} & \textbf{5.23} \\
\midrule

\multirow{2}{*}{Q3 235B}
& EAGLE-3    & 2.55$\times$ & 3.55 & 3.22$\times$ & 4.70 & 3.46$\times$ & 5.05 & 2.73$\times$ & 3.50 & 2.78$\times$ & 3.90 & 2.95$\times$ & 4.14 \\
& MARS      & \textbf{3.12$\times$} & \textbf{4.65} & \textbf{3.67$\times$} & \textbf{5.55} & \textbf{3.88$\times$} & \textbf{5.95} & \textbf{3.25$\times$} & \textbf{4.70} & \textbf{3.20$\times$} & \textbf{4.85} & \textbf{3.42$\times$} & \textbf{5.14} \\
\bottomrule
\end{tabular}
\caption{\textbf{Overall performance.} Speedup ratios and average acceptance lengths $\tau$ of different methods under non-greedy decoding (temperature$=1$). V denotes Vicuna, L31 denotes LLaMA-Instruct 3.1, and Q3 denotes Qwen3. SpS denotes standard speculative sampling with Vicuna-68M as the draft model. Because of the characteristics of SD, once all drafts are accepted, the sequence will be appended with a bonus token generated by the target model. Therefore, the theoretical upper limit of $\tau$ is \textit{K}+1, which is 8.}
\label{tab:Overall}
\end{table*}

\subsection{Experimental Setup}

\newcommand{\method}{\textsc{MARS}}

In this section, we describe the models, datasets, evaluation metrics, baselines, and implementation details.

\paragraph{Models.}
We evaluate \method\ on state-of-the-art open-source LLMs that cover both chat-oriented and reasoning-oriented settings,
including Vicuna-13B v1.3 \citep{vicuna2023},
LLaMA-3.1-8B-Instruct, LLaMA-3.1-70B-Instruct \citep{grattafiori2024llama3herdmodels},
Qwen3-8B, Qwen3-32B, and Qwen3-235B-A22B \citep{qwen3technicalreport}.
All models listed above are used as the \textbf{target models} to be accelerated.
For each target model, we additionally employ the corresponding \textbf{EAGLE-3} draft model trained for that target to generate speculative drafts.
 \method\ is applied \emph{without} modifying the target model parameters.
Due to hardware constraints, we do not include experiments on models larger than \textbf{235}B parameters.

\paragraph{Datasets.}
Following prior work on lossless LLM decoding acceleration \citep{li2024eagle2fasterinferencelanguage, li2025eagle3scalinginferenceacceleration},
we evaluate on a diverse set of tasks that are commonly used for chat, code generation, reasoning, instruction following, translation, and summarization:
MT-Bench \citep{zheng2023judgingllmasajudgemtbenchchatbot},
HumanEval \citep{chen2021codex}
and MBPP \citep{austin2021program},
GSM8K \citep{cobbe2021gsm8k},
AlpacaEval/Alpaca \citep{alpaca},
WMT19 \citep{wmt19translate}, 
and CNN/DailyMail \citep{hermann2015teaching}.

\paragraph{Evaluation Metrics.}
Different from strictly lossless acceleration, our method is a \emph{lossy} variant of speculative decoding (SD),which may deviate from the target model's exact token distribution.
Therefore, we evaluate \method\ along two axes: \textbf{generation quality} and \textbf{efficiency}.
Specifically, we report:

\begin{itemize}\setlength\itemsep{0.2em}
    \item \textbf{Accuracy}: task-specific accuracy to quantify the quality impact introduced by lossy SD.
    For classification-style benchmarks, we report exact-match accuracy.
    For reasoning/QA benchmarks (e.g., GSM8K), we follow standard exact-match accuracy over final answers
    \citep{cobbe2021gsm8k}.
    For code-generation benchmarks (e.g., HumanEval), we report avg@4 as the primary accuracy-oriented metric
    \citep{chen2021codex}.
    \item \textbf{Speedup Ratio}: end-to-end decoding speed relative to vanilla autoregressive decoding,
    measured under the same hardware, decoding parameters, and stopping criteria.
    % \item \textbf{Throughput}: tokens per second measured at \textbf{[batch size(s)]} to reflect serving efficiency.
    \item \textbf{Average Acceptance Length $\tau$}: the average number of tokens committed per draft--verify cycle,
    capturing how many draft tokens are used per verification step (larger is generally more efficient).
\end{itemize}

% We do not include \textit{Acceptance Rate} in our metrics to keep the evaluation consistent across different
% verification/commitment policies and to focus on end-to-end efficiency and task-level correctness.

\paragraph{Comparison.}
We use vanilla autoregressive decoding as the primary baseline (Speedup $=1.00\times$).
We compare \method\ with representative lossless (or commonly used) decoding-acceleration methods,
including standard speculative sampling \citep{leviathan2023fast, chen2023speculative},
Prompt Lookup Decoding \citep{somasundaram2024pldacceleratingllminference},
Medusa \citep{cai2024medusa},
Lookahead decoding \citep{fu2024breaksequentialdependencyllm},
EAGLE-2 \citep{li2024eagle2fasterinferencelanguage},
and EAGLE-3 \citep{ li2025eagle3scalinginferenceacceleration}.
For all baselines, we follow the officially released implementations and default hyperparameters when available and otherwise match decoding settings (e.g., temperature, max length, and stopping criteria) to ensure fairness.
\paragraph{Implementation Details.} For a fair comparison in our performance evaluation, we fix the draft length to 7 for all methods. In the draft process, we maintain Topk = 10 during the draft tree building, and no additional pruning is performed during the draft stage, keeping the same settings as Eagle3.
Since our study primarily targets sampling scenarios with temperature $> 0$, we do not report results at temperature $= 0$.
All experiments are conducted on NVIDIA H100 80GB GPUs.
Models with $\leq 32$B parameters are benchmarked on a single GPU, while models with $> 32$B parameters are evaluated using 8 GPUs.

% Table~\ref{tab:Overall} reports overall decoding efficiency at temperature$=1$. Across all model families and scales, our method consistently improves over EAGLE-3, delivering higher speedup and longer average accepted lengths $\tau$.

% On \textbf{Vicuna-13B}, we achieve 3.74$\times$ mean speedup with $\tau=7.20$, compared to 3.12$\times$ and $\tau=5.64$ for EAGLE-3. The gains are also consistent on \textbf{L31 8B} (4.00$\times$ vs.\ 3.24$\times$) and \textbf{L33 70B} (4.76$\times$ vs.\ 4.40$\times$), with $\tau$ increased in all cases. We observe similar improvements on \textbf{Qwen3} models: 3.23$\times$ (vs.\ 2.96$\times$) on \textbf{Q3 8B}, 3.49$\times$ (vs.\ 3.01$\times$) on \textbf{Q3 32B}, and 3.42$\times$ (vs.\ 2.95$\times$) on \textbf{Q3 235B}. 

% Overall, the consistent increase in $\tau$ indicates better draft efficiency under stochastic decoding, translating into reliable end-to-end speedups.

\subsection{Overall Performance}
\label{sec:overall}

Table~\ref{tab:Overall} summarizes overall decoding efficiency at temperature$=1$. Across all model families and scales, our method consistently improves over EAGLE-3, yielding higher speedup ratios and longer average accepted lengths $\tau$.

On \textbf{Vicuna-13B}, our method achieves a 3.74$\times$ mean speedup with $\tau=7.20$, compared to 3.12$\times$ and $\tau=5.64$ for EAGLE-3. The improvements remain consistent on \textbf{LLaMA-Instruct 3.1-8B} (4.00$\times$ vs.\ 3.24$\times$) and \textbf{LLaMA-Instruct 3.1-70B} (4.76$\times$ vs.\ 4.40$\times$), with $\tau$ increasing in all cases. We observe similar gains on \textbf{Qwen3-8B} (3.23$\times$ vs.\ 2.96$\times$), \textbf{Qwen3-32B} (3.49$\times$ vs.\ 3.01$\times$), and \textbf{Qwen3-235B-A22B} (3.42$\times$ vs.\ 2.95$\times$). 

Overall, the consistent increase in $\tau$ indicates better draft efficiency under stochastic decoding, translating into reliable end-to-end acceleration.

Although our experiments focus on sampling ($T{>}0$), MARS is equally applicable to greedy decoding ($T{=}0$). We provide $T{=}0$ results in Appendix~\ref{app:greedy}: MARS remains consistently faster than EAGLE-3 while preserving comparable task accuracy.

% \input{latex/figures/recovery}

% \subsection{Accuracy Preservation}
% Since our approach is a lossy speculative decoding (SD) method, accelerated decoding is not guaranteed to be exactly
% identical to standard decoding. We therefore evaluate \textbf{accuracy preservation} by comparing accelerated outputs
% against the same target model under the same decoding configuration, and report task accuracy together with a
% \textbf{recovery ratio} (accelerated accuracy normalized by the baseline).

% As shown in Figure~\ref{fig:rec}, our method achieves near-complete recovery across tasks and model scales
% (98.1--100\%), with \textbf{100\%} recovery on HumanEval and 99.2--100\% on MBPP.
% We observe negligible degradation on code benchmarks, likely because occasional divergences are mostly superficial
% (e.g., naming/formatting) and rarely affect functional correctness.

\subsection{Ablation Studies}
\label{ablation}
In this section, we quantify the contribution of key hyper-parameters in \method\ and analyze the
quality--efficiency trade-off under a unified evaluation setup.
Unlike strictly lossless speculative decoding, our approach is a \emph{lossy} SD variant, so we report both
\textbf{task performance} and \textbf{decoding efficiency}.
Concretely, we measure (i) \textbf{Accuracy} on each benchmark,
(ii) \textbf{Speedup} (end-to-end latency reduction relative to vanilla autoregressive decoding),
and (iii) \textbf{Precision}, which captures the agreement/consistency of the generated output with the vanilla
autoregressive output under the same decoding configuration.

% \paragraph{Logit ratio threshold $\theta$.}
% \label{sec:ablation_theta}
% \input{latex/figures/acc_speed}
% The logit ratio threshold $\theta$ controls the degree of \emph{adaptive relaxation} in \method.
% Concretely, when the draft token $\hat{v}_i$ does not match the target model’s top-1 token $v^{(1)}$ but matches the top-2
% token $v^{(2)}$, \method\ relaxes the verification rule and still commits $\hat{v}_i$ if the logit ratio
% $r_i = z_{i,(2)} / z_{i,(1)}$ exceeds $\theta$ (i.e., the top-1/top-2 margin is small).
% Therefore, a smaller $\theta$ makes the policy more permissive by triggering relaxation more often, which increases speedup
% but may introduce additional deviations that slightly reduce accuracy.
% In contrast, a larger $\theta$ makes relaxation harder to trigger (more conservative verification), leading to fewer relaxed
% acceptances and thus lower speedup, with accuracy typically saturating rather than improving further.
% As shown in Figure~\ref{fig:accspeed}, speedup decreases monotonically as $\theta$ increases, while accuracy consistently peaks
% around $\theta \approx 0.90$ across both HumanEval and GSM8K with $K\in\{7,10\}$.
% We therefore use $\theta=\textbf{0.90}$ as the default for a favorable accuracy--speed trade-off.

\paragraph{Logit ratio threshold $\theta$.}
\label{sec:ablation_theta}
\begin{figure}[t] \centering
    \includegraphics[width=\linewidth]{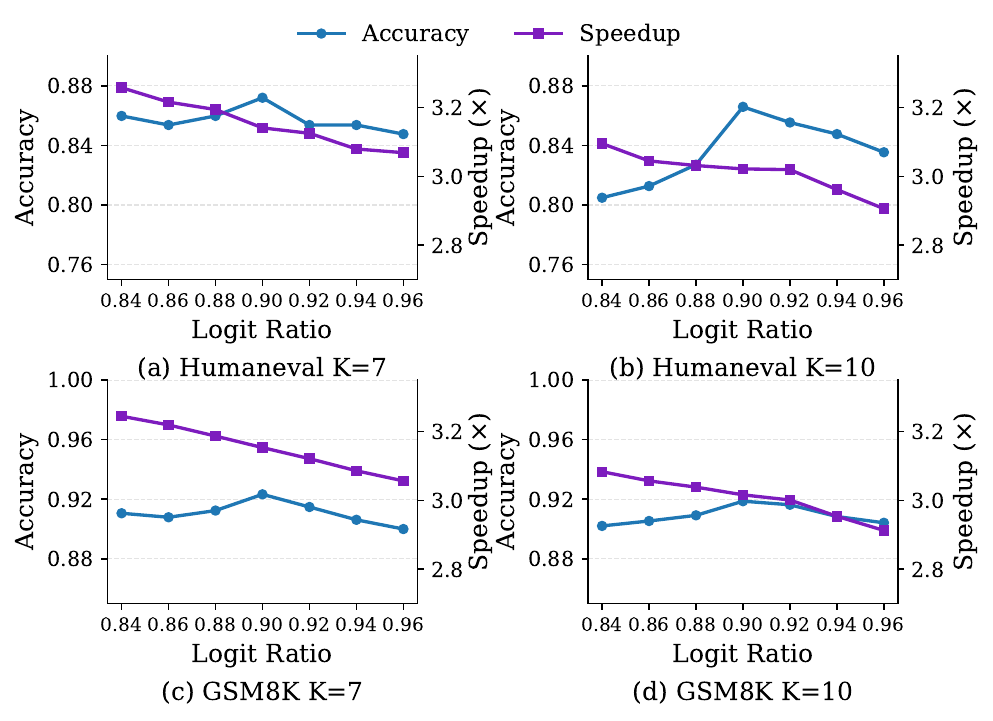}
    \caption{\textbf{Effect of the logit ratio threshold $\theta$ on the quality-efficiency trade-off.}
We sweep over $\theta \in \{0.84, 0.86, 0.88, 0.90, 0.92, 0.94, 0.96\}$ and report accuracy (blue, left axis) and speedup (purple, right axis) on HumanEval and GSM8K with $K\in\{7,10\}$ and $T{=}1$. Increasing $\theta$ consistently reduces speedup, while accuracy typically peaks around $\theta\approx0.90$, indicating a balanced default choice.}
    \label{fig:accspeed}
\end{figure}
The logit ratio threshold $\theta$ controls the degree of \emph{adaptive
relaxation} in \method. Concretely, when the draft token $\hat{v}_i$ does not
match the target model's top-1 token $v^{(1)}$ but matches the top-2 token
$v^{(2)}$, \method\ relaxes the verification rule and still commits
$\hat{v}_i$ if the logit ratio $r_i = z_{i,(2)} / z_{i,(1)}$ exceeds $\theta$
(i.e., the top-1/top-2 margin is small). A smaller $\theta$ makes the policy
more permissive, increasing speedup, but potentially introducing additional
deviations; a larger $\theta$ yields more conservative verification with lower
speedup.
 
As shown in Figure~\ref{fig:accspeed}, speedup decreases monotonically as
$\theta$ increases, while accuracy consistently peaks around
$\theta \approx 0.90$ across both HumanEval and GSM8K with
$K\in\{7,10\}$ (all at $T{=}1$, Qwen3-8B). Notably, accuracy is
\emph{not} monotonically increasing with $\theta$ under sampling. This is
because overly strict verification (high $\theta$) triggers frequent
rollbacks, forcing the model to re-sample from the target's full distribution.
Under $T{=}1$, this resampling can introduce stochastic tail noise that
offsets the benefit of stricter verification. The WMT19 results in
Table~\ref{tab:wmt19} corroborate this pattern: on large-vocabulary models
(Qwen3-8B/32B, $\sim$152K tokens), quality peaks near $\theta{=}0.90$ and
fluctuates at higher values, whereas on Vicuna-13B ($\sim$32K tokens) the
trend is more monotonic, suggesting that the effect is amplified by
vocabulary size. We therefore use $\theta=\textbf{0.90}$ as the default for a
favorable accuracy-speed trade-off.

\begin{table*}[t]
\centering
% 使用 resizebox 确保表格不会溢出页面宽度
\resizebox{\textwidth}{!}{
\begin{tabular}{ll ccc ccc ccc ccc ccc}
\toprule
 & & \multicolumn{15}{c}{Temperature} \\
\cmidrule(lr){3-17}
 & & \multicolumn{3}{c}{0.2} & \multicolumn{3}{c}{0.4} & \multicolumn{3}{c}{0.6} & \multicolumn{3}{c}{0.8} & \multicolumn{3}{c}{1} \\
\cmidrule(lr){3-5} \cmidrule(lr){6-8} \cmidrule(lr){9-11} \cmidrule(lr){12-14} \cmidrule(lr){15-17}
Dataset & \textit{K} & Speedup & $\tau$ & Acc & Speedup & $\tau$ & Acc & Speedup & $\tau$ & Acc & Speedup & $\tau$ & Acc & Speedup & $\tau$ & Acc \\
\midrule
\multirow{5}{*}{GSM8K} 
& baseline & 1.0x & \textbackslash & 0.934 & 1.0x & \textbackslash & 0.932 & 1.0x & \textbackslash & 0.909 & 1.0x & \textbackslash & 0.901 & 1.0x & \textbackslash & 0.857 \\

& 6  & 3.96x & 5.79 & 0.929 & 3.89x & 5.75 & 0.927 & 3.97x & 5.76 & 0.894 & 3.95x & 5.76 & 0.896 & 3.97x & 5.74 & 0.851 \\
& 9  & 4.15x & 6.58 & 0.921 & 4.14x & 6.60 & 0.927 & 4.17x & 6.62 & 0.893 & 4.23x & 6.60 & 0.896 & 4.19x & 6.52 & 0.893 \\
& 12 & 4.09x & 6.87 & 0.921 & 4.07x & 6.85 & 0.924 & 4.13x & 6.83 & 0.904 & 4.05x & 6.80 & 0.904 & 4.07x & 6.76 & 0.916 \\
& 15 & 3.94x & 6.92 & 0.933 & 4.00x & 7.00 & 0.933 & 3.95x & 6.93 & 0.907 & 3.91x & 6.90 & 0.909 & 3.90x & 6.85 & 0.917 \\

\midrule
\multirow{5}{*}{Humaneval} 
& baseline
& 1.0x & \textbackslash & 0.896
& 1.0x & \textbackslash & 0.878
& 1.0x & \textbackslash & 0.878
& 1.0x & \textbackslash & 0.866
& 1.0x & \textbackslash & 0.854 \\ 

& 6
& 3.72x & 5.35 & 0.896
& 3.65x & 5.34 & 0.890
& 3.65x & 5.35 & 0.878
& 3.63x & 5.31 & 0.878
& 3.70x & 5.31 & 0.848 \\

& 9
& 3.84x & 5.98 & 0.884
& 3.86x & 5.94 & 0.860
& 3.84x & 5.93 & 0.860
& 3.83x & 5.93 & 0.842
& 3.90x & 6.03 & 0.866 \\

& 12
& 3.78x & 6.12 & 0.878
& 3.70x & 6.12 & 0.884
& 3.74x & 6.06 & 0.872
& 3.79x & 6.13 & 0.860
& 3.78x & 6.08 & 0.872 \\

& 15
& 3.58x & 6.18 & 0.878
& 3.59x & 6.15 & 0.878
& 3.61x & 6.08 & 0.854
& 3.63x & 6.14 & 0.860
& 3.59x & 6.14 & 0.848 \\ 
\bottomrule
\end{tabular}
}
\caption{
Ablation results on \textbf{Qwen3-32B} over \textbf{GSM8K} and \textbf{HumanEval}, studying the effects of
\textbf{temperature} ($t$) and \textbf{draft length} ($K$) in MARS.
For each dataset, we report \textbf{Speedup} (end-to-end decoding speed relative to vanilla autoregressive decoding),
$\boldsymbol{\tau}$ (average committed tokens per draft--verify cycle), and \textbf{Acc} (task accuracy).
``baseline'' denotes vanilla autoregressive decoding under the same temperature.
}
\label{tab:abla}
\end{table*}

\paragraph{Temperature $t$.}
Temperature $t$ reshapes the sampling distribution and thus affects both generation quality and SD behavior.
As shown in Table~\ref{tab:abla}, \textbf{Speedup} and $\boldsymbol{\tau}$ are largely stable across temperatures
($t\in\{0.2,0.4,0.6,0.8,1.0\}$) on both GSM8K and HumanEval, while \textbf{accuracy consistently decreases as $t$ increases}
(for both the AR baseline and our method).
Overall, temperature do not affect the generation quality of our method within acceptable limits. Stable efficiency improvements are achieved under different temperature settings.

\paragraph{Draft length $K$.}
Table~\ref{tab:abla} shows that increasing $K$ leads to larger $\tau$ (more tokens committed per draft-verify cycle),
but \textbf{speedup is not monotonic in $K$}.
Moderate draft lengths (often $K{=}9$) achieve the best acceleration, whereas overly large $K$ can slightly reduce speedup,
suggesting additional drafter/verification overhead.
Accuracy varies with $K$ and task type (more sensitive on HumanEval than GSM8K), indicating a quality--efficiency trade-off.
We treat these results as a \emph{sensitivity analysis} and do not select hyperparameters based on this study.

\subsection{Quality Preservation across Tasks and Granularities}
\label{sec:quality_preserve}
 
Since MARS is a lossy variant of speculative decoding, we evaluate holistic
generation quality on MT-Bench (GPT-5 as judge,
$T{=}1$) across five models and eight categories; MARS preserves the
baseline's average score within $\pm$0.18 points on all models (see
Appendix~\ref{app:mtbench} for full results).
 
\paragraph{Segment-Level Fidelity (CNN/DailyMail).}
To measure whether relaxation alters the structural overlap between generated
and reference texts, we report ROUGE-L on
CNN/DailyMail (zero-shot, $K{=}7$,
temperature$=1$). As shown in Table~\ref{tab:cnndm}, MARS scores are within
0.001--0.002 of the baseline across all five models (average: 0.1935 vs.\
0.1943), well within the range of stochastic decoding variance. This indicates
that the longest common subsequence structure of the generated summaries is
effectively unchanged.
% ---------- CNN/DailyMail ROUGE-L ----------
\begin{table}[t]
\centering
\small
\setlength{\tabcolsep}{4pt}
\begin{tabular}{lcccccc}
\toprule
Method & L31 8B & L31 70B & Q3 8B & Q3 32B & V 13B \\
\midrule
Baseline & 0.1699 & 0.1827 & 0.1972 & 0.1944 & 0.2272 \\
EAGLE-3  & 0.1707 & 0.1812 & 0.1985 & 0.1948 & 0.2260 \\
MARS     & 0.1692 & 0.1812 & 0.1966 & 0.1945 & 0.2260 \\
\bottomrule
\end{tabular}
\caption{Segment-level quality preservation on CNN/DailyMail (ROUGE-L, zero-shot, $\theta{=}0.9$, $K{=}7$, $T{=}1$). For all models, the difference between the average MARS scores and the baseline is 0.0008.}
\label{tab:cnndm}
\end{table}
 
\paragraph{Token-Level Fidelity (WMT19).}
We further evaluate on WMT19 Zh-En translation using BLEU and chrF($\beta{=}2$), both of which are sensitive to token and character-level deviations. Table~\ref{tab:wmt19} reports results across a range of thresholds $\theta \in \{0.84, \ldots, 0.98\}$. At the default $\theta{=}0.9$, MARS closely tracks the baseline on all models: for example, Qwen3-8B achieves 29.67 BLEU (baseline: 29.71) and 59.81 chrF (baseline: 60.25). Quality is well preserved at $\theta \ge 0.90$, while more aggressive relaxation ($\theta < 0.88$) can introduce measurable degradation, particularly on larger-vocabulary models.

% ---------- WMT19 BLEU / chrF ----------
\begin{table}[t]
\centering
\small
\setlength{\tabcolsep}{4pt}
\begin{tabular}{lcccccc}
\toprule
 & \multicolumn{2}{c}{Qwen3-8B} & \multicolumn{2}{c}{Qwen3-32B} & \multicolumn{2}{c}{Vicuna-13B} \\
\cmidrule(lr){2-3} \cmidrule(lr){4-5} \cmidrule(lr){6-7}
Setting & BLEU & chrF & BLEU & chrF & BLEU & chrF \\
\midrule
Baseline       & 29.71 & 60.25 & 30.01 & 60.47 & 22.62 & 53.84 \\
EAGLE-3        & 29.99 & 60.40 & 30.02 & 60.35 & 22.26 & 53.66 \\
\midrule
$\theta=0.84$  & 28.70 & 58.88 & 26.28 & 58.61 & 20.33 & 52.32 \\
$\theta=0.86$  & 29.12 & 58.97 & 26.98 & 59.01 & 20.40 & 52.56 \\
$\theta=0.88$  & 29.34 & 59.12 & 29.08 & 59.53 & 21.21 & 52.42 \\
$\theta=0.90$  & 29.67 & 59.81 & 29.76 & 60.10 & 21.50 & 53.19 \\
$\theta=0.92$  & 29.59 & 59.46 & 29.62 & 60.10 & 21.58 & 53.14 \\
$\theta=0.94$  & 29.10 & 59.61 & 29.57 & 60.18 & 21.95 & 53.35 \\
$\theta=0.96$  & 29.39 & 59.88 & 28.29 & 60.07 & 21.90 & 53.09 \\
$\theta=0.98$  & 29.22 & 59.78 & 28.57 & 60.24 & 21.97 & 53.73 \\
\bottomrule
\end{tabular}
\caption{Token-level quality on WMT19 Zh-En (BLEU / chrF($\beta{=}2$), $K{=}7$, $T{=}1$) under varying logit ratio thresholds. Baseline and EAGLE-3 are shown for reference. Quality is well preserved at $\theta \ge 0.90$; aggressive relaxation ($\theta < 0.88$) degrades notably.}
\label{tab:wmt19}
\end{table}

\subsection{Framework-Decoupled Verification}
\label{sec:framework_decoupled}

MARS modifies only the verification accept/reject rule and is therefore
decoupled from the drafting architecture. To verify this, we integrate MARS
into \textbf{Standard Speculative Decoding (SPD)}, keeping its draft$\to$verify pipeline entirely
unchanged and only replacing the verification criterion. We evaluate two
draft$\to$target pairs—Qwen3-0.6B$\to$Qwen3-32B and
LLaMA-3.1-8B$\to$LLaMA-3.1-70B—on GSM8K, HumanEval, and WMT19
($T{=}1$, $\gamma{=}6$).

As shown in Table~\ref{tab:spd}, MARS consistently increases the average
accepted length $\tau$ over vanilla SPD on both model pairs, translating
into higher end-to-end speedups, and task accuracy is preserved. These results
confirm that MARS is not tied to EAGLE-3 and provides consistent gains as a
plug-and-play verification strategy across different speculative decoding
frameworks.

\begin{table*}[t]
\centering
\small
\setlength{\tabcolsep}{3.5pt}
\begin{tabular}{llccccccccc}
\toprule
 &  & \multicolumn{3}{c}{GSM8K} & \multicolumn{3}{c}{HumanEval} & \multicolumn{3}{c}{WMT19} \\
\cmidrule(lr){3-5} \cmidrule(lr){6-8} \cmidrule(lr){9-11}
Model & Method & Speedup & $\tau$ & Acc & Speedup & $\tau$ & Acc & Speedup & $\tau$ & BLEU \\
\midrule
\multirow{3}{*}{\shortstack[l]{Qwen3\\0.6B$\to$32B}}
 & Baseline & 1$\times$ & -- & 0.901 & 1$\times$ & -- & 0.896 & 1$\times$ & -- & 29.13 \\
 & SPD      & 2.15$\times$ & 4.23 & 0.893 & 2.23$\times$ & 4.33 & 0.896 & 1.51$\times$ & 2.92 & 29.16 \\
 & SPD+MARS & 2.70$\times$ & 4.94 & 0.901 & 2.76$\times$ & 4.90 & 0.902 & 1.85$\times$ & 3.44 & 29.27 \\
\midrule
\multirow{3}{*}{\shortstack[l]{LLaMA3\\8B$\to$70B}}
 & Baseline & 1$\times$ & -- & 0.942 & 1$\times$ & -- & 0.730 & 1$\times$ & -- & 27.96 \\
 & SPD      & 3.05$\times$ & 5.07 & 0.934 & 3.15$\times$ & 5.73 & 0.712 & 2.20$\times$ & 4.16 & 27.84 \\
 & SPD+MARS & 3.81$\times$ & 5.88 & 0.934 & 4.36$\times$ & 6.37 & 0.732 & 2.55$\times$ & 4.59 & 27.80 \\
\bottomrule
\end{tabular}
\caption{Integration of MARS into Standard Speculative Decoding (SPD) ($T{=}1$, $\gamma{=}6$). MARS increases $\tau$ and speedup over vanilla SPD on both draft$\to$target pairs while preserving or improving task accuracy, confirming its framework-decoupled nature.}
\label{tab:spd}
\end{table*}

\section{Related work}

Speculative Decoding (SD) accelerates LLM inference by verifying cheap draft tokens in parallel, a paradigm established by \cite{leviathan2023fast, chen2023speculative}. Current research enhances SD through three key dimensions: model alignment, verification strategies, and lossy trade-offs.

\subsection{Alignment of Draft and Target Models}
The acceptance rate $\alpha$ relies heavily on the distributional alignment between models. To minimize divergence, \textbf{Knowledge Distillation} methods like DistillSpec \cite{zhou2024distillspecimprovingspeculativedecoding} and SSD \cite{spector2023acceleratingllminferencestaged} fine-tune drafters to mimic the target, while \cite{liu2024onlinespeculativedecoding} updates the drafter online using rejected tokens. 
Recently, \textbf{Architectural Alignment} has proven more effective than independent models. Medusa \cite{cai2024medusa} and Hydra \cite{ankner2024hydra} append decoding heads to the target backbone, whereas EAGLE \cite{li2025eaglespeculativesamplingrequires, li2024eagle2fasterinferencelanguage} utilizes the target's feature history for context-aware drafting. Alternatively, REST \cite{he2024restretrievalbasedspeculativedecoding} aligns generation with corpus statistics via retrieval-based drafting.

\subsection{Better Verification Strategies}
Standard verification often under-utilizes GPU parallelism. \textbf{Tree-based Verification} addresses this by evaluating token trees instead of chains. SpecInfer \cite{miao2024specinfer} and SpecTr \cite{sun2024spectrfastspeculativedecoding} leverage tree attention masks to verify multiple branches in a single pass, with Sequoia \cite{chen2025sequoiascalablerobusthardwareaware} further optimizing tree structures dynamically.
Distinctly, \textbf{Draft-Free} strategies operate without proxy models. Jacobi Decoding \cite{leviathan2023fast} employs fixed-point iteration for multi-token prediction. Similarly, Lookahead Decoding \cite{fu2024breaksequentialdependencyllm} utilizes parallel n-gram generation within the target model to break sequential dependencies efficiently.
\vspace{-0.2cm}

\subsection{Lossy Speculative Decoding}

Recent research advances from lossless verification to \textit{lossy} SD, trading negligible quality degradation for significant latency reduction. One primary direction relaxes verification criteria based on model confidence. \citet{kim2023speculativedecodingbiglittle} proposed bypassing target model verification for high-confidence draft tokens, while \citet{zhang2025draft} introduced adaptive thresholds tuned by context entropy to dynamically balance coherence and speed. 

% Alternatively, other works shift from exact token matching to semantic consistency. \citet{gong2024semantic} utilized embedding similarity to accept synonymous draft tokens, effectively increasing acceptance rates. Furthermore, \citet{liu2024lossy} developed a fuzzy verification framework using quantized models, which approximates the target distribution to minimize verification overhead.

% Strict adherence to the target distribution limits efficiency. Lossy SD relaxes verification criteria to trade negligible quality drops for speed. \textbf{Confidence-based} methods like BiLD \cite{kim2023speculativedecodingbiglittle} bypass verification when draft confidence exceeds a threshold. \textbf{Semantic-based} approaches \cite{zhang2023bi, du2024acc} accept tokens based on embedding similarity rather than exact matching. Furthermore, \cite{ye2024collaborative} proposes collaborative decoding to blend logits from both models, smoothing the rejection boundary for open-ended generation.

\section{Conclusion}
In this paper, we identify a fundamental inefficiency in speculative decoding: strict exact-match verification incurs substantial overhead in low-margin regimes where the target model exhibits weak decisiveness.
To address this issue, we propose \method, a training-free and plug-and-play verification strategy that adaptively adjusts acceptance rigor based on the target model’s logit margin.
By accepting plausible runner-up draft tokens in \textbf{low-margin regimes}, \method\ significantly reduces unnecessary rollbacks while preserving generation quality.
Extensive experiments across diverse model families (8B to 235B) show that our approach consistently outperforms state-of-the-art baselines, achieving a maximum speedup of \textbf{4.76$\times$} on LLaMA-3.1-70B.
Our results suggest that replacing rigid token matching with margin-aware adaptive verification is a critical step toward more efficient LLM inference.

\section{Limitations}
MARS focuses on adapting the verification rule based on local decision margins derived from the target model’s logits. While we demonstrate that a single global threshold performs consistently across a wide range of models and tasks, more fine-grained or context-dependent adaptation strategies may further improve robustness. In addition, our method operates at the token level and is designed to be fully plug-and-play with existing speculative decoding frameworks; exploring margin-aware decisions at higher semantic or structural levels is beyond the scope of this work.

% \section*{Acknowledgments} 

\bibliography{custom}

@inproceedings{miao2024specinfer,
  title={SpecInfer: Accelerating Generative Large Language Model Serving with Tree-based Speculative Inference and Verification},
  author={Miao, Xupeng and Oliaro, Gabriele and He, Zhihao and Schulze, Aaron and Parisot, Cheng-Hao and Paszke, Adam and Jia, Zhihao},
  booktitle={Proceedings of the 29th ACM International Conference on Architectural Support for Programming Languages and Operating Systems, Volume 2 (ASPLOS '24)},
  year={2024}
}

@misc{sun2024spectrfastspeculativedecoding,
      title={SpecTr: Fast Speculative Decoding via Optimal Transport}, 
      author={Ziteng Sun and Ananda Theertha Suresh and Jae Hun Ro and Ahmad Beirami and Himanshu Jain and Felix Yu},
      year={2024},
      eprint={2310.15141},
      archivePrefix={arXiv},
      primaryClass={cs.LG},
      url={https://arxiv.org/abs/2310.15141}, 
}

@misc{zhou2024distillspecimprovingspeculativedecoding,
      title={DistillSpec: Improving Speculative Decoding via Knowledge Distillation}, 
      author={Yongchao Zhou and Kaifeng Lyu and Ankit Singh Rawat and Aditya Krishna Menon and Afshin Rostamizadeh and Sanjiv Kumar and Jean-François Kagy and Rishabh Agarwal},
      year={2024},
      eprint={2310.08461},
      archivePrefix={arXiv},
      primaryClass={cs.CL},
      url={https://arxiv.org/abs/2310.08461}, 
}

@misc{spector2023acceleratingllminferencestaged,
      title={Accelerating LLM Inference with Staged Speculative Decoding}, 
      author={Benjamin Spector and Chris Re},
      year={2023},
      eprint={2308.04623},
      archivePrefix={arXiv},
      primaryClass={cs.AI},
      url={https://arxiv.org/abs/2308.04623}, 
}

@misc{li2025eaglespeculativesamplingrequires,
      title={EAGLE: Speculative Sampling Requires Rethinking Feature Uncertainty}, 
      author={Yuhui Li and Fangyun Wei and Chao Zhang and Hongyang Zhang},
      year={2025},
      eprint={2401.15077},
      archivePrefix={arXiv},
      primaryClass={cs.LG},
      url={https://arxiv.org/abs/2401.15077}, 
}

@misc{li2024eagle2fasterinferencelanguage,
      title={EAGLE-2: Faster Inference of Language Models with Dynamic Draft Trees}, 
      author={Yuhui Li and Fangyun Wei and Chao Zhang and Hongyang Zhang},
      year={2024},
      eprint={2406.16858},
      archivePrefix={arXiv},
      primaryClass={cs.CL},
      url={https://arxiv.org/abs/2406.16858}, 
}

@misc{he2024restretrievalbasedspeculativedecoding,
      title={REST: Retrieval-Based Speculative Decoding}, 
      author={Zhenyu He and Zexuan Zhong and Tianle Cai and Jason D. Lee and Di He},
      year={2024},
      eprint={2311.08252},
      archivePrefix={arXiv},
      primaryClass={cs.CL},
      url={https://arxiv.org/abs/2311.08252}, 
}

@misc{chen2025sequoiascalablerobusthardwareaware,
      title={Sequoia: Scalable, Robust, and Hardware-aware Speculative Decoding}, 
      author={Zhuoming Chen and Avner May and Ruslan Svirschevski and Yuhsun Huang and Max Ryabinin and Zhihao Jia and Beidi Chen},
      year={2025},
      eprint={2402.12374},
      archivePrefix={arXiv},
      primaryClass={cs.CL},
      url={https://arxiv.org/abs/2402.12374}, 
}

@misc{fu2024breaksequentialdependencyllm,
      title={Break the Sequential Dependency of LLM Inference Using Lookahead Decoding}, 
      author={Yichao Fu and Peter Bailis and Ion Stoica and Hao Zhang},
      year={2024},
      eprint={2402.02057},
      archivePrefix={arXiv},
      primaryClass={cs.LG},
      url={https://arxiv.org/abs/2402.02057}, 
}

@misc{kim2023speculativedecodingbiglittle,
      title={Speculative Decoding with Big Little Decoder}, 
      author={Sehoon Kim and Karttikeya Mangalam and Suhong Moon and Jitendra Malik and Michael W. Mahoney and Amir Gholami and Kurt Keutzer},
      year={2023},
      eprint={2302.07863},
      archivePrefix={arXiv},
      primaryClass={cs.CL},
      url={https://arxiv.org/abs/2302.07863}, 
}

@misc{liu2024onlinespeculativedecoding,
      title={Online Speculative Decoding}, 
      author={Xiaoxuan Liu and Lanxiang Hu and Peter Bailis and Alvin Cheung and Zhijie Deng and Ion Stoica and Hao Zhang},
      year={2024},
      eprint={2310.07177},
      archivePrefix={arXiv},
      primaryClass={cs.AI},
      url={https://arxiv.org/abs/2310.07177}, 
}

@misc{vicuna2023,
    title = {Vicuna: An Open-Source Chatbot Impressing GPT-4 with 90\%* ChatGPT Quality},
    url = {https://lmsys.org/blog/2023-03-30-vicuna/},
    author = {Chiang, Wei-Lin and Li, Zhuohan and Lin, Zi and Sheng, Ying and Wu, Zhanghao and Zhang, Hao and Zheng, Lianmin and Zhuang, Siyuan and Zhuang, Yonghao and Gonzalez, Joseph E. and Stoica, Ion and Xing, Eric P.},
    month = {March},
    year = {2023}
}

@misc{grattafiori2024llama3herdmodels,
      title={The Llama 3 Herd of Models}, 
      author={Aaron Grattafiori and Abhimanyu Dubey and Abhinav Jauhri and Abhinav Pandey and Abhishek Kadian and Ahmad Al-Dahle and Aiesha Letman and Akhil Mathur and Alan Schelten and Alex Vaughan and Amy Yang and Angela Fan and Anirudh Goyal and Anthony Hartshorn and Aobo Yang and Archi Mitra and Archie Sravankumar and Artem Korenev and Arthur Hinsvark and Arun Rao and Aston Zhang and Aurelien Rodriguez and Austen Gregerson and Ava Spataru and Baptiste Roziere and Bethany Biron and Binh Tang and Bobbie Chern and Charlotte Caucheteux and Chaya Nayak and Chloe Bi and Chris Marra and Chris McConnell and Christian Keller and Christophe Touret and Chunyang Wu and Corinne Wong and Cristian Canton Ferrer and Cyrus Nikolaidis and Damien Allonsius and Daniel Song and Danielle Pintz and Danny Livshits and Danny Wyatt and David Esiobu and Dhruv Choudhary and Dhruv Mahajan and Diego Garcia-Olano and Diego Perino and Dieuwke Hupkes and Egor Lakomkin and Ehab AlBadawy and Elina Lobanova and Emily Dinan and Eric Michael Smith and Filip Radenovic and Francisco Guzmán and Frank Zhang and Gabriel Synnaeve and Gabrielle Lee and Georgia Lewis Anderson and Govind Thattai and Graeme Nail and Gregoire Mialon and Guan Pang and Guillem Cucurell and Hailey Nguyen and Hannah Korevaar and Hu Xu and Hugo Touvron and Iliyan Zarov and Imanol Arrieta Ibarra and Isabel Kloumann and Ishan Misra and Ivan Evtimov and Jack Zhang and Jade Copet and Jaewon Lee and Jan Geffert and Jana Vranes and Jason Park and Jay Mahadeokar and Jeet Shah and Jelmer van der Linde and Jennifer Billock and Jenny Hong and Jenya Lee and Jeremy Fu and Jianfeng Chi and Jianyu Huang and Jiawen Liu and Jie Wang and Jiecao Yu and Joanna Bitton and Joe Spisak and Jongsoo Park and Joseph Rocca and Joshua Johnstun and Joshua Saxe and Junteng Jia and Kalyan Vasuden Alwala and Karthik Prasad and Kartikeya Upasani and Kate Plawiak and Ke Li and Kenneth Heafield and Kevin Stone and Khalid El-Arini and Krithika Iyer and Kshitiz Malik and Kuenley Chiu and Kunal Bhalla and Kushal Lakhotia and Lauren Rantala-Yeary and Laurens van der Maaten and Lawrence Chen and Liang Tan and Liz Jenkins and Louis Martin and Lovish Madaan and Lubo Malo and Lukas Blecher and Lukas Landzaat and Luke de Oliveira and Madeline Muzzi and Mahesh Pasupuleti and Mannat Singh and Manohar Paluri and Marcin Kardas and Maria Tsimpoukelli and Mathew Oldham and Mathieu Rita and Maya Pavlova and Melanie Kambadur and Mike Lewis and Min Si and Mitesh Kumar Singh and Mona Hassan and Naman Goyal and Narjes Torabi and Nikolay Bashlykov and Nikolay Bogoychev and Niladri Chatterji and Ning Zhang and Olivier Duchenne and Onur Çelebi and Patrick Alrassy and Pengchuan Zhang and Pengwei Li and Petar Vasic and Peter Weng and Prajjwal Bhargava and Pratik Dubal and Praveen Krishnan and Punit Singh Koura and Puxin Xu and Qing He and Qingxiao Dong and Ragavan Srinivasan and Raj Ganapathy and Ramon Calderer and Ricardo Silveira Cabral and Robert Stojnic and Roberta Raileanu and Rohan Maheswari and Rohit Girdhar and Rohit Patel and Romain Sauvestre and Ronnie Polidoro and Roshan Sumbaly and Ross Taylor and Ruan Silva and Rui Hou and Rui Wang and Saghar Hosseini and Sahana Chennabasappa and Sanjay Singh and Sean Bell and Seohyun Sonia Kim and Sergey Edunov and Shaoliang Nie and Sharan Narang and Sharath Raparthy and Sheng Shen and Shengye Wan and Shruti Bhosale and Shun Zhang and Simon Vandenhende and Soumya Batra and Spencer Whitman and Sten Sootla and Stephane Collot and Suchin Gururangan and Sydney Borodinsky and Tamar Herman and Tara Fowler and Tarek Sheasha and Thomas Georgiou and Thomas Scialom and Tobias Speckbacher and Todor Mihaylov and Tong Xiao and Ujjwal Karn and Vedanuj Goswami and Vibhor Gupta and Vignesh Ramanathan and Viktor Kerkez and Vincent Gonguet and Virginie Do and Vish Vogeti and Vítor Albiero and Vladan Petrovic and Weiwei Chu and Wenhan Xiong and Wenyin Fu and Whitney Meers and Xavier Martinet and Xiaodong Wang and Xiaofang Wang and Xiaoqing Ellen Tan and Xide Xia and Xinfeng Xie and Xuchao Jia and Xuewei Wang and Yaelle Goldschlag and Yashesh Gaur and Yasmine Babaei and Yi Wen and Yiwen Song and Yuchen Zhang and Yue Li and Yuning Mao and Zacharie Delpierre Coudert and Zheng Yan and Zhengxing Chen and Zoe Papakipos and Aaditya Singh and Aayushi Srivastava and Abha Jain and Adam Kelsey and Adam Shajnfeld and Adithya Gangidi and Adolfo Victoria and Ahuva Goldstand and Ajay Menon and Ajay Sharma and Alex Boesenberg and Alexei Baevski and Allie Feinstein and Amanda Kallet and Amit Sangani and Amos Teo and Anam Yunus and Andrei Lupu and Andres Alvarado and Andrew Caples and Andrew Gu and Andrew Ho and Andrew Poulton and Andrew Ryan and Ankit Ramchandani and Annie Dong and Annie Franco and Anuj Goyal and Aparajita Saraf and Arkabandhu Chowdhury and Ashley Gabriel and Ashwin Bharambe and Assaf Eisenman and Azadeh Yazdan and Beau James and Ben Maurer and Benjamin Leonhardi and Bernie Huang and Beth Loyd and Beto De Paola and Bhargavi Paranjape and Bing Liu and Bo Wu and Boyu Ni and Braden Hancock and Bram Wasti and Brandon Spence and Brani Stojkovic and Brian Gamido and Britt Montalvo and Carl Parker and Carly Burton and Catalina Mejia and Ce Liu and Changhan Wang and Changkyu Kim and Chao Zhou and Chester Hu and Ching-Hsiang Chu and Chris Cai and Chris Tindal and Christoph Feichtenhofer and Cynthia Gao and Damon Civin and Dana Beaty and Daniel Kreymer and Daniel Li and David Adkins and David Xu and Davide Testuggine and Delia David and Devi Parikh and Diana Liskovich and Didem Foss and Dingkang Wang and Duc Le and Dustin Holland and Edward Dowling and Eissa Jamil and Elaine Montgomery and Eleonora Presani and Emily Hahn and Emily Wood and Eric-Tuan Le and Erik Brinkman and Esteban Arcaute and Evan Dunbar and Evan Smothers and Fei Sun and Felix Kreuk and Feng Tian and Filippos Kokkinos and Firat Ozgenel and Francesco Caggioni and Frank Kanayet and Frank Seide and Gabriela Medina Florez and Gabriella Schwarz and Gada Badeer and Georgia Swee and Gil Halpern and Grant Herman and Grigory Sizov and Guangyi and Zhang and Guna Lakshminarayanan and Hakan Inan and Hamid Shojanazeri and Han Zou and Hannah Wang and Hanwen Zha and Haroun Habeeb and Harrison Rudolph and Helen Suk and Henry Aspegren and Hunter Goldman and Hongyuan Zhan and Ibrahim Damlaj and Igor Molybog and Igor Tufanov and Ilias Leontiadis and Irina-Elena Veliche and Itai Gat and Jake Weissman and James Geboski and James Kohli and Janice Lam and Japhet Asher and Jean-Baptiste Gaya and Jeff Marcus and Jeff Tang and Jennifer Chan and Jenny Zhen and Jeremy Reizenstein and Jeremy Teboul and Jessica Zhong and Jian Jin and Jingyi Yang and Joe Cummings and Jon Carvill and Jon Shepard and Jonathan McPhie and Jonathan Torres and Josh Ginsburg and Junjie Wang and Kai Wu and Kam Hou U and Karan Saxena and Kartikay Khandelwal and Katayoun Zand and Kathy Matosich and Kaushik Veeraraghavan and Kelly Michelena and Keqian Li and Kiran Jagadeesh and Kun Huang and Kunal Chawla and Kyle Huang and Lailin Chen and Lakshya Garg and Lavender A and Leandro Silva and Lee Bell and Lei Zhang and Liangpeng Guo and Licheng Yu and Liron Moshkovich and Luca Wehrstedt and Madian Khabsa and Manav Avalani and Manish Bhatt and Martynas Mankus and Matan Hasson and Matthew Lennie and Matthias Reso and Maxim Groshev and Maxim Naumov and Maya Lathi and Meghan Keneally and Miao Liu and Michael L. Seltzer and Michal Valko and Michelle Restrepo and Mihir Patel and Mik Vyatskov and Mikayel Samvelyan and Mike Clark and Mike Macey and Mike Wang and Miquel Jubert Hermoso and Mo Metanat and Mohammad Rastegari and Munish Bansal and Nandhini Santhanam and Natascha Parks and Natasha White and Navyata Bawa and Nayan Singhal and Nick Egebo and Nicolas Usunier and Nikhil Mehta and Nikolay Pavlovich Laptev and Ning Dong and Norman Cheng and Oleg Chernoguz and Olivia Hart and Omkar Salpekar and Ozlem Kalinli and Parkin Kent and Parth Parekh and Paul Saab and Pavan Balaji and Pedro Rittner and Philip Bontrager and Pierre Roux and Piotr Dollar and Polina Zvyagina and Prashant Ratanchandani and Pritish Yuvraj and Qian Liang and Rachad Alao and Rachel Rodriguez and Rafi Ayub and Raghotham Murthy and Raghu Nayani and Rahul Mitra and Rangaprabhu Parthasarathy and Raymond Li and Rebekkah Hogan and Robin Battey and Rocky Wang and Russ Howes and Ruty Rinott and Sachin Mehta and Sachin Siby and Sai Jayesh Bondu and Samyak Datta and Sara Chugh and Sara Hunt and Sargun Dhillon and Sasha Sidorov and Satadru Pan and Saurabh Mahajan and Saurabh Verma and Seiji Yamamoto and Sharadh Ramaswamy and Shaun Lindsay and Shaun Lindsay and Sheng Feng and Shenghao Lin and Shengxin Cindy Zha and Shishir Patil and Shiva Shankar and Shuqiang Zhang and Shuqiang Zhang and Sinong Wang and Sneha Agarwal and Soji Sajuyigbe and Soumith Chintala and Stephanie Max and Stephen Chen and Steve Kehoe and Steve Satterfield and Sudarshan Govindaprasad and Sumit Gupta and Summer Deng and Sungmin Cho and Sunny Virk and Suraj Subramanian and Sy Choudhury and Sydney Goldman and Tal Remez and Tamar Glaser and Tamara Best and Thilo Koehler and Thomas Robinson and Tianhe Li and Tianjun Zhang and Tim Matthews and Timothy Chou and Tzook Shaked and Varun Vontimitta and Victoria Ajayi and Victoria Montanez and Vijai Mohan and Vinay Satish Kumar and Vishal Mangla and Vlad Ionescu and Vlad Poenaru and Vlad Tiberiu Mihailescu and Vladimir Ivanov and Wei Li and Wenchen Wang and Wenwen Jiang and Wes Bouaziz and Will Constable and Xiaocheng Tang and Xiaojian Wu and Xiaolan Wang and Xilun Wu and Xinbo Gao and Yaniv Kleinman and Yanjun Chen and Ye Hu and Ye Jia and Ye Qi and Yenda Li and Yilin Zhang and Ying Zhang and Yossi Adi and Youngjin Nam and Yu and Wang and Yu Zhao and Yuchen Hao and Yundi Qian and Yunlu Li and Yuzi He and Zach Rait and Zachary DeVito and Zef Rosnbrick and Zhaoduo Wen and Zhenyu Yang and Zhiwei Zhao and Zhiyu Ma},
      year={2024},
      eprint={2407.21783},
      archivePrefix={arXiv},
      primaryClass={cs.AI},
      url={https://arxiv.org/abs/2407.21783}, 
}

@misc{qwen3technicalreport,
      title={Qwen3 Technical Report}, 
      author={Qwen Team},
      year={2025},
      eprint={2505.09388},
      archivePrefix={arXiv},
      primaryClass={cs.CL},
      url={https://arxiv.org/abs/2505.09388}, 
}

@misc{li2025eagle3scalinginferenceacceleration,
      title={EAGLE-3: Scaling up Inference Acceleration of Large Language Models via Training-Time Test}, 
      author={Yuhui Li and Fangyun Wei and Chao Zhang and Hongyang Zhang},
      year={2025},
      eprint={2503.01840},
      archivePrefix={arXiv},
      primaryClass={cs.CL},
      url={https://arxiv.org/abs/2503.01840}, 
}

@misc{zheng2023judgingllmasajudgemtbenchchatbot,
      title={Judging LLM-as-a-Judge with MT-Bench and Chatbot Arena}, 
      author={Lianmin Zheng and Wei-Lin Chiang and Ying Sheng and Siyuan Zhuang and Zhanghao Wu and Yonghao Zhuang and Zi Lin and Zhuohan Li and Dacheng Li and Eric P. Xing and Hao Zhang and Joseph E. Gonzalez and Ion Stoica},
      year={2023},
      eprint={2306.05685},
      archivePrefix={arXiv},
      primaryClass={cs.CL},
      url={https://arxiv.org/abs/2306.05685}, 
}

@article{chen2021codex,
  title={Evaluating Large Language Models Trained on Code},
  author={Mark Chen and Jerry Tworek and Heewoo Jun and Qiming Yuan and Henrique Ponde de Oliveira Pinto and Jared Kaplan and Harri Edwards and Yuri Burda and Nicholas Joseph and Greg Brockman and Alex Ray and Raul Puri and Gretchen Krueger and Michael Petrov and Heidy Khlaaf and Girish Sastry and Pamela Mishkin and Brooke Chan and Scott Gray and Nick Ryder and Mikhail Pavlov and Alethea Power and Lukasz Kaiser and Mohammad Bavarian and Clemens Winter and Philippe Tillet and Felipe Petroski Such and Dave Cummings and Matthias Plappert and Fotios Chantzis and Elizabeth Barnes and Ariel Herbert-Voss and William Hebgen Guss and Alex Nichol and Alex Paino and Nikolas Tezak and Jie Tang and Igor Babuschkin and Suchir Balaji and Shantanu Jain and William Saunders and Christopher Hesse and Andrew N. Carr and Jan Leike and Josh Achiam and Vedant Misra and Evan Morikawa and Alec Radford and Matthew Knight and Miles Brundage and Mira Murati and Katie Mayer and Peter Welinder and Bob McGrew and Dario Amodei and Sam McCandlish and Ilya Sutskever and Wojciech Zaremba},
  year={2021},
  eprint={2107.03374},
  archivePrefix={arXiv},
  primaryClass={cs.LG}
}

@article{cobbe2021gsm8k,
  title={Training Verifiers to Solve Math Word Problems},
  author={Cobbe, Karl and Kosaraju, Vineet and Bavarian, Mohammad and Chen, Mark and Jun, Heewoo and Kaiser, Lukasz and Plappert, Matthias and Tworek, Jerry and Hilton, Jacob and Nakano, Reiichiro and Hesse, Christopher and Schulman, John},
  journal={arXiv preprint arXiv:2110.14168},
  year={2021}
}

@misc{alpaca,
  author = {Rohan Taori and Ishaan Gulrajani and Tianyi Zhang and Yann Dubois and Xuechen Li and Carlos Guestrin and Percy Liang and Tatsunori B. Hashimoto },
  title = {Stanford Alpaca: An Instruction-following LLaMA model},
  year = {2023},
  publisher = {GitHub},
  journal = {GitHub repository},
  howpublished = {\url{https://github.com/tatsu-lab/stanford_alpaca}},
}

@inproceedings{hermann2015teaching,
  title     = {Teaching Machines to Read and Comprehend},
  author    = {Hermann, Karl Moritz and Kocisk\'{y}, Tom\'{a}\v{s} and Grefenstette, Edward and Espeholt, Lasse and Kay, Will and Suleyman, Mustafa and Blunsom, Phil},
  booktitle = {Advances in Neural Information Processing Systems},
  pages     = {1693--1701},
  year      = {2015}
}

@misc{somasundaram2024pldacceleratingllminference,
      title={PLD+: Accelerating LLM inference by leveraging Language Model Artifacts}, 
      author={Shwetha Somasundaram and Anirudh Phukan and Apoorv Saxena},
      year={2024},
      eprint={2412.01447},
      archivePrefix={arXiv},
      primaryClass={cs.CL},
      url={https://arxiv.org/abs/2412.01447}, 
}

@article{chen2023speculative,
  title        = {Accelerating Large Language Model Decoding with Speculative Sampling},
  author       = {Chen, Charlie and Borgeaud, Sebastian and Irving, Geoffrey and Lespiau, Jean-Baptiste and Sifre, Laurent and Jumper, John},
  journal      = {arXiv preprint arXiv:2302.01318},
  year         = {2023}
}

@article{cai2024medusa,
  title        = {Medusa: Simple LLM Inference Acceleration Framework with Multiple Decoding Heads},
  author       = {Cai, Tianle and Li, Yuhong and Geng, Zhengyang and Peng, Hongwu and Lee, Jason D. and Chen, Deming and Dao, Tri},
  journal      = {arXiv preprint arXiv:2401.10774},
  year         = {2024}
}

@article{ankner2024hydra,
  title        = {Hydra: Sequentially-Dependent Draft Heads for Medusa Decoding},
  author       = {Ankner, Zachary and Parthasarathy, Rishab and Nrusimha, Aniruddha and Rinard, Christopher and Ragan-Kelley, Jonathan and Brandon, William},
  journal      = {arXiv preprint arXiv:2402.05109},
  year         = {2024}
}

@article{shazeer2019fast,
  title        = {Fast Transformer Decoding: One Write-Head is All You Need},
  author       = {Shazeer, Noam},
  journal      = {arXiv preprint arXiv:1911.02150},
  year         = {2019}
}

@article{li2025loosely,
  title   = {Training-Free Loosely Speculative Decoding: Accepting Semantically Correct Drafts Beyond Exact Match},
  author  = {Li, Jinze and Xu, Yixing and Li, Guanchen and others},
  journal = {arXiv preprint arXiv:2511.22972},
  year    = {2025}
}

@inproceedings{zhang2025draft,
  title     = {Draft Model Knows When to Stop: Self-Verification Speculative Decoding for Long-Form Generation},
  author    = {Zhang, Ziyin and Xu, Jiahao and Liang, Tian and Chen, Xingyu and He, Zhiwei and Wang, Rui and Tu, Zhaopeng},
  booktitle = {Proceedings of the 2025 Conference on Empirical Methods in Natural Language Processing},
  pages     = {16696--16708},
  year      = {2025},
  address   = {Suzhou, China},
  publisher = {Association for Computational Linguistics},
  doi       = {10.18653/v1/2025.emnlp-main.844},
  url       = {https://aclanthology.org/2025.emnlp-main.844}
}

@inproceedings{bachmann2025judge,
  title={Judge decoding: Faster speculative sampling requires going beyond model alignment},
  author={Bachmann, Gregor and Alistarh, Dan},
  booktitle={The Thirteenth International Conference on Learning Representations (ICLR)},
  year={2025},
  note={Oral Presentation}
}

@article{austin2021program,
  title={Program Synthesis with Large Language Models},
  author={Austin, Jacob and Odena, Augustus and Nye, Maxwell and Bosma, Maarten and Michalewski, Henryk and Dohan, David and Jiang, Ellen and Cai, Carrie and Terry, Michael and Le, Quoc and others},
  journal={arXiv preprint arXiv:2108.07732},
  year={2021}
}

@misc{leviathan2023fast,
      title={Fast Inference from Transformers via Speculative Decoding}, 
      author={Yaniv Leviathan and Matan Kalman and Yossi Matias},
      year={2023},
      eprint={2211.17192},
      archivePrefix={arXiv},
      primaryClass={cs.LG},
      url={https://arxiv.org/abs/2211.17192}, 
}

@ONLINE {wmt19translate,
    author = "Wikimedia Foundation",
    title  = "ACL 2019 Fourth Conference on Machine Translation (WMT19), Shared Task: Machine Translation of News",
    year={2019},
    url    = "http://www.statmt.org/wmt19/translation-task.html"
}
% \newpage
\appendix

% \appendix
% \section{Theoretical Properties of the Ratio Indicator}

% \subsection{Scaling Robustness}

% We provide a simple theoretical interpretation for the ratio-based indicator.
% Assume that the target logits at decoding step $t$ can be expressed as
% \begin{equation}
% \mathbf{z}_t = s_t \, \tilde{\mathbf{z}}_t,
% \end{equation}
% where $s_t > 0$ is an unknown scalar that may vary across decoding steps, and
% $\tilde{\mathbf{z}}_t$ represents the underlying normalized scores.

% \begin{proposition}
% Under the scaling model above, the logit ratio
% $r_t = z_2 / z_1$ is invariant to the scaling factor $s_t$, whereas the fixed logit
% margin $\Delta_t = z_1 - z_2$ scales linearly with $s_t$.
% \end{proposition}

% \begin{proof}
% By substitution, we have $z_1 = s_t \tilde{z}_1$ and $z_2 = s_t \tilde{z}_2$, hence
% \begin{equation}
% r_t = \frac{z_2}{z_1} = \frac{s_t \tilde{z}_2}{s_t \tilde{z}_1}
% = \frac{\tilde{z}_2}{\tilde{z}_1},
% \end{equation}
% which is independent of $s_t$.
% In contrast, the margin satisfies
% \begin{equation}
% \Delta_t = z_1 - z_2 = s_t(\tilde{z}_1 - \tilde{z}_2),
% \end{equation}
% which scales linearly with $s_t$.
% \end{proof}

% \subsection{Local Regret Interpretation}

% Selecting $x^{(2)}$ instead of $x^{(1)}$ incurs a local negative log-likelihood
% difference
% \begin{equation}
% \log \frac{P(x^{(1)} \mid x_{<t})}{P(x^{(2)} \mid x_{<t})} = \Delta_t.
% \end{equation}
% As $r_t \to 1$, we have $\Delta_t \to 0$, implying that the local regret of choosing
% $x^{(2)}$ approaches zero.
% This provides a justification for ratio-guided tie-breaking in low-margin regimes.

\section{Additional Analysis}
\begin{figure*}[t] \centering
    \includegraphics[width=1\textwidth]{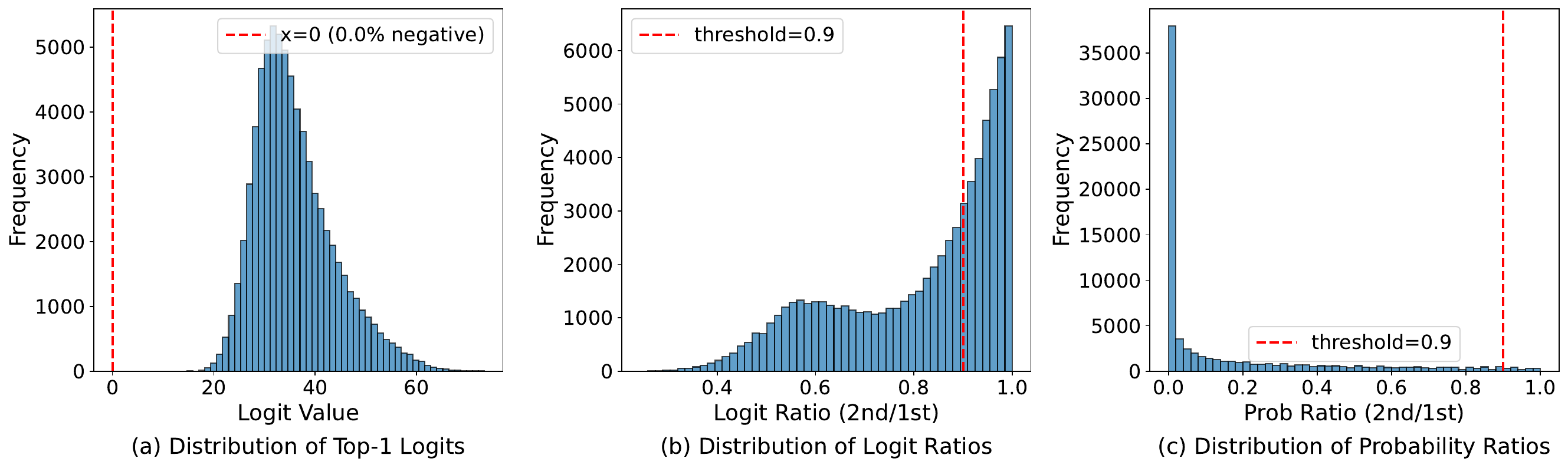}
    \caption{
Histograms of top-2 statistics on \textbf{Qwen3-8B}.
(a) Distribution of the top-1 logit values; the red dashed line marks $x{=}0$ (0.0\% of top-1 logits are negative).
(b) Distribution of logit ratios between the 2nd and 1st candidates ($z_2/z_1$); the red dashed line indicates the threshold ($0.9$).
(c) Distribution of probability ratios between the 2nd and 1st candidates ($p_2/p_1$); the red dashed line indicates the same threshold ($0.9$).
}
    \label{fig:his}
\end{figure*}
This appendix provides supplementary analyses for understanding the behavior of our relaxation criterion.
Figure~\ref{fig:his} reports the empirical distributions of (a) top-1 logit values, (b) logit ratios between the
top-2 candidates ($z_2/z_1$), and (c) probability ratios between the top-2 candidates ($p_2/p_1$) on \textbf{Qwen3-8B}.
We mark key reference points with red dashed lines: $x{=}0$ in (a), and a fixed threshold of $0.9$ in (b--c).

\section{Greedy Decoding Results}
\label{app:greedy}

Although our experiments primarily target sampling settings ($T{>}0$), MARS is fully compatible with greedy decoding ($T{=}0$). We report results with $K{=}7$ in Table~\ref{tab:greedy}. Across all three model families, MARS consistently achieves higher speedup and longer average acceptance length $\tau$ than EAGLE-3 while maintaining comparable task accuracy. We note that under greedy decoding, both EAGLE-3 and MARS may exhibit minor accuracy deviations from the baseline due to numerical differences in tree attention and KV cache management; this is a known artifact of speculative decoding implementations rather than a property of the verification rule itself.

\begin{table*}[h]
\centering
\resizebox{\linewidth}{!}{%
\begin{tabular}{ll ccc ccc ccc}
\toprule
 & & \multicolumn{3}{c}{\textbf{GSM8K}} & \multicolumn{3}{c}{\textbf{HumanEval}} & \multicolumn{3}{c}{\textbf{Avg}} \\
\cmidrule(lr){3-5} \cmidrule(lr){6-8} \cmidrule(lr){9-11}
\textbf{Model} & \textbf{Method} & Speedup & $\tau$ & Acc & Speedup & $\tau$ & Acc & Speedup & $\tau$ & Acc \\
\midrule
\multirow{3}{*}{Qwen3-8B}
 & Baseline & 1.00$\times$ & -- & 0.967 & 1.00$\times$ & -- & 0.835 & 1.00$\times$ & -- & 0.901 \\
 & EAGLE-3  & 2.83$\times$ & 5.20 & 0.959 & 2.53$\times$ & 4.87 & 0.848 & 2.68$\times$ & 5.03 & 0.903 \\
 & MARS     & 3.13$\times$ & 5.93 & 0.950 & 2.69$\times$ & 5.48 & 0.854 & 2.91$\times$ & 5.71 & 0.902 \\
\midrule
\multirow{3}{*}{Vicuna-13B}
 & Baseline & 1.00$\times$ & -- & 0.223 & 1.00$\times$ & -- & 0.184 & 1.00$\times$ & -- & 0.204 \\
 & EAGLE-3  & 3.93$\times$ & 6.35 & 0.207 & 4.30$\times$ & 7.27 & 0.190 & 4.12$\times$ & 6.81 & 0.198 \\
 & MARS     & 4.13$\times$ & 7.18 & 0.215 & 4.61$\times$ & 7.73 & 0.183 & 4.37$\times$ & 7.46 & 0.199 \\
\midrule
\multirow{3}{*}{LLaMA-3.1-8B}
 & Baseline & 1.00$\times$ & -- & 0.893 & 1.00$\times$ & -- & 0.677 & 1.00$\times$ & -- & 0.785 \\
 & EAGLE-3  & 3.57$\times$ & 5.86 & 0.901 & 4.10$\times$ & 6.42 & 0.677 & 3.93$\times$ & 6.14 & 0.789 \\
 & MARS     & 3.89$\times$ & 6.69 & 0.884 & 4.37$\times$ & 6.93 & 0.671 & 4.13$\times$ & 6.81 & 0.778 \\
\bottomrule
\end{tabular}%
}
\caption{Performance under greedy decoding ($T{=}0$, $K{=}7$). Speedup is relative to vanilla autoregressive decoding. $\tau$ denotes the average acceptance length per draft-verify cycle. Acc reports task-specific accuracy (exact-match for GSM8K, avg@4 for HumanEval).}
\label{tab:greedy}
\end{table*}

\section{MT-Bench Fine-Grained Results}
\label{app:mtbench}

To validate generation quality beyond math and coding tasks, we evaluate on MT-Bench across eight diverse categories using GPT-5 as the judge (single-turn, $T{=}1$). As shown in Table~\ref{tab:mtbench}, MARS preserves the baseline's generation quality across all categories and model scales, with average scores within $\pm$0.1 of the baseline in most cases.

\begin{table*}[h]
\centering
\resizebox{\linewidth}{!}{%
\begin{tabular}{ll cccccccc c}
\toprule
\textbf{Model} & \textbf{Method} & \textbf{Coding} & \textbf{Extraction} & \textbf{Humanities} & \textbf{Math} & \textbf{Reasoning} & \textbf{Roleplay} & \textbf{STEM} & \textbf{Writing} & \textbf{Avg} \\
\midrule
\multirow{3}{*}{Vicuna-13B}
 & Baseline & 2.56 & 4.22 & 4.40 & 2.90 & 4.11 & 3.80 & 4.10 & 4.90 & 3.87 \\
 & EAGLE-3  & 2.60 & 3.89 & 4.50 & 2.70 & 4.11 & 3.70 & 3.90 & 5.20 & 3.83 \\
 & MARS     & 2.60 & 4.22 & 4.50 & 2.80 & 4.11 & 3.80 & 3.90 & 5.20 & 3.89 \\
\midrule
\multirow{3}{*}{Qwen3-8B}
 & Baseline & 6.22 & 7.78 & 6.20 & 10.00 & 7.67 & 6.30 & 7.78 & 7.30 & 7.41 \\
 & EAGLE-3  & 6.11 & 7.78 & 6.50 & 10.00 & 7.78 & 6.50 & 7.67 & 7.10 & 7.43 \\
 & MARS     & 7.00 & 8.11 & 6.40 & 10.00 & 7.78 & 6.30 & 7.90 & 7.20 & 7.59 \\
\midrule
\multirow{3}{*}{Qwen3-32B}
 & Baseline & 8.50 & 8.44 & 6.50 & 9.30 & 8.78 & 6.90 & 7.30 & 8.00 & 7.97 \\
 & EAGLE-3  & 8.71 & 8.33 & 6.80 & 9.20 & 8.70 & 7.00 & 7.70 & 7.90 & 8.04 \\
 & MARS     & 9.29 & 8.44 & 6.70 & 9.30 & 8.56 & 6.90 & 7.30 & 8.10 & 8.07 \\
\midrule
\multirow{3}{*}{LLaMA-3.1-8B}
 & Baseline & 4.12 & 6.33 & 4.50 & 6.80 & 6.56 & 5.00 & 4.70 & 6.10 & 5.51 \\
 & EAGLE-3  & 4.71 & 6.22 & 4.20 & 6.80 & 6.67 & 5.10 & 4.70 & 6.40 & 5.60 \\
 & MARS     & 4.43 & 6.22 & 4.10 & 6.80 & 6.56 & 5.40 & 5.00 & 6.10 & 5.58 \\
\midrule
\multirow{3}{*}{LLaMA-3.1-70B}
 & Baseline & 6.71 & 7.89 & 5.00 & 9.90 & 8.44 & 6.90 & 5.80 & 7.00 & 7.21 \\
 & EAGLE-3  & 6.43 & 8.11 & 5.30 & 9.90 & 8.44 & 7.20 & 5.60 & 6.70 & 7.21 \\
 & MARS     & 6.25 & 8.00 & 5.10 & 9.90 & 8.33 & 6.90 & 5.60 & 6.90 & 7.12 \\
\bottomrule
\end{tabular}%
}
\caption{MT-Bench single-turn scores by category (GPT-5 judge, $T{=}1$). Higher is better. MARS shows no systematic quality degradation across diverse generation tasks.}
\label{tab:mtbench}
\end{table*}

\end{document}